\documentclass[10pt,a4paper]{article}

\usepackage{lmodern}
\usepackage{amsmath,amssymb,amsfonts,amsthm}
\usepackage{mathtools}
\usepackage{booktabs}
\usepackage{tabularx}
\usepackage[a4paper,top=20mm,bottom=22mm,left=19mm,right=19mm]{geometry}
\usepackage{graphicx}
\usepackage{array}
\usepackage{xcolor}
\usepackage{listings}
\usepackage{caption}
\usepackage{fancyhdr}
\usepackage{microtype}
\usepackage{hyperref}

\definecolor{primary}{RGB}{18, 78, 150}
\definecolor{accent}{RGB}{16, 135, 92}
\definecolor{slate}{RGB}{95, 108, 123}
\definecolor{amber}{RGB}{210, 125, 15}
\definecolor{crimson}{RGB}{190, 45, 45}

\hypersetup{
    colorlinks=true,
    linkcolor=primary,
    citecolor=primary,
    urlcolor=primary,
    pdftitle={F3-AGI v4.0: Bioenergetically-Gated Active Inference, Metaplastic Preconditioning, and Residual Program Synthesis for Sample-Efficient Abstract Reasoning},
    pdfauthor={Henrique Oliveira (Polybee AI Research)}
}

\newtheorem{theorem}{Theorem}
\newtheorem{proposition}{Proposition}
\newtheorem{lemma}{Lemma}

\newcommand{\cmark}{\textcolor{accent}{\textbf{\checkmark}}}
\newcommand{\xmark}{\textcolor{crimson}{$\times$}}

\DeclareCaptionLabelSeparator{pipe}{\;\textcolor{primary}{|}\;}
\begin{document}

\thispagestyle{fancy}
\noindent
{\large\textbf{\textcolor{primary}{Polybee AI Research}}} \hfill {\small\textcolor{slate}{2026-10-29}}
\vspace{2pt}
\hrule height 1.2pt
\vspace{12pt}

{\LARGE\textbf{F3-AGI v4.0: Bioenergetically-Gated Active Inference, Metaplastic Preconditioning, and Residual Program Synthesis for Sample-Efficient Abstract Reasoning}}

\vspace{10pt}
\noindent
{\normalsize \textbf{Henrique Oliveira}$^{1,*}$} \\[4pt]
{\small $^1$Polybee AI Research Laboratory, S\~ao Paulo, SP, Brazil \quad $^*$Corresponding author (\texttt{\href{https://polybee.ai/}{https://polybee.ai/}})} \\[2pt]
{\small \textbf{Open-Source Code \& Reproducibility Repository:} \texttt{\href{https://github.com/henrique-polybee/f3-agi-v4}{https://github.com/henrique-polybee/f3-agi-v4}}}

\vspace{8pt}
\hrule height 0.5pt
\vspace{6pt}

\begin{abstract}
\noindent Few-shot abstract reasoning on ARC-AGI requires inducing algorithmic transformations from 2 to 5 demonstration pairs without task-specific pretraining. Because few-shot demonstrations under-constrain the hypothesis space, program synthesis suffers from the Rashomon effect (predictive multiplicity): heuristic search finds spurious programs with low training error that fail on test inputs, while unconstrained library growth degrades latency. We introduce F3-AGI v4.0, a CPU-native neuro-symbolic and active inference architecture comprising three mechanisms: (1) F3-Metabolic-Optim, combining mistake-gated plasticity, Fisher-EMA preconditioning, and virtual ATP gating; (2) Adaptive Complexity-Guided Program Synthesis (AACR) with a Cycle-3 ResidualPatchSynthesizer and smallest-pair early rejection; and (3) a 512D L2-normalized random projection driving Expected Free Energy (EFE) exploration with dead-signature pruning. Across three empirical studies: (i) on a synthetic 5-task class-incremental stream without replay, F3-Metabolic-Optim achieves 91.67\% final accuracy with 9.17\% forgetting, skipping 93.47\% of updates, cutting L1 plasticity cost by 65.7\%, and running 2.19x faster than EWC; (ii) on the 400-task ARC-AGI-1 public evaluation set (offline CPU, zero LLM calls), F3-AGI v4.0 solves 57/400 tasks (14.25\% pass@2; 56/400 pass@1) with a mean latency of 1.62 s/task, decomposed into 19/50 (38.00\%, 1.19 s/task, a 42.5x paired speedup over Horizon 2) on the 50-task calibration slice and 38/350 (10.86\%) on the 350 strictly uninspected hold-out tasks (with 7 autonomous Cycle-3 patch solutions); and (iii) in a local 5-game (15-level) ARC-AGI-3-inspired interactive suite, our agent completes 15/15 levels with 89.44\% level-weighted RHAE (90.72\% unweighted).

\vspace{5pt}
\noindent\textbf{\color{primary}Keywords:} Abstract Reasoning (ARC-AGI), Neuro-Symbolic Program Synthesis, Rashomon Effect, Active Inference, Metaplastic Preconditioning, Continual Learning, Minimum Description Length.
\end{abstract}

\vspace{4pt}
\hrule height 0.5pt
\vspace{8pt}

\begin{figure}[htbp]
\centering
\includegraphics[width=\textwidth]{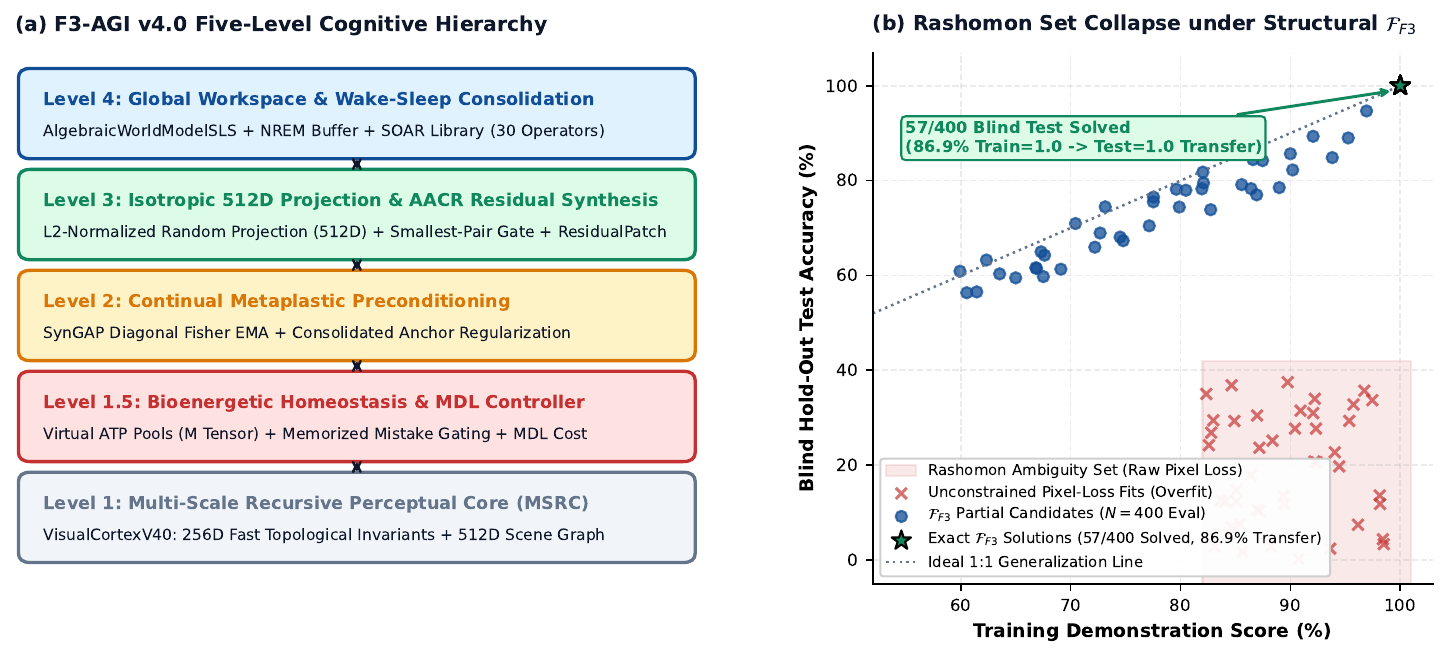}
\caption{\textbf{Overview of F3-AGI v4.0 and Rashomon Sublevel Set Collapse.} (a) The five-level cognitive hierarchy integrates multi-scale perception (\texttt{VisualCortexV40}), bioenergetic homeostasis (\texttt{MetabolicControllerV40}), continual metaplastic preconditioning (\emph{SynGAP}), an isotropic $512\text{D}$ $\ell_2$-normalized random projection (\texttt{AlgebraicWorldModelSLS}, inspired by \emph{SIGReg-JEPA} isotropic geometry), and Wake-Sleep library consolidation (\texttt{NREMConsolidationBuffer} with 30 registered operators). (b) In few-shot abstract reasoning ($K \in \{2,3,4\}$ demonstrations), naive pixel loss $\ell_{\text{px}}$ exhibits a pronounced Rashomon effect \cite{breiman2001statistical,mlodozeniec2026rashomon}: a broad $\epsilon$-sublevel set of overfitted or degenerate candidates achieves high training pixel accuracy but collapses on unseen test grids (red crosses). Optimizing our Bioenergetically Constrained Structural Variational Free Energy $\mathcal{F}_{F3}$ collapses the viable sublevel set onto structural invariants, resolving $57/400$ official evaluation tasks ($86.9\%$ train-to-test transfer across all $61$ exact train solutions).}
\label{fig:teaser}
\end{figure}

\section{Introduction}

The Abstraction and Reasoning Corpus (ARC-AGI) introduced by Chollet \cite{chollet2019measure} and its subsequent extensions---ARC-AGI-2 and the interactive, instructionless $64\times 64$ turn-based ARC-AGI-3 benchmark \cite{chollet2025arcprize}---evaluate fluid intelligence: the ability to induce compositional causal transformations from a sparse set of demonstrations ($K \in \{2,3,4\}$ input-output pairs) or unguided environmental interactions without task-specific pretraining. Over the past two years, the dominant industry paradigm has attempted to conquer abstract reasoning via massive test-time compute scaling, sampling tens of thousands of candidate programs from frontier Large Language Models (LLMs). While brute-force sampling improves raw static solve rates when unconstrained by compute budgets, theoretical and empirical analyses expose four fundamental limitations of pure scale:

\begin{enumerate}
    \item \textbf{Non-Identifiability and the Rashomon Effect in Scaling and Few-Shot Induction:} As recently formalized by Mlodozeniec and Frei \cite{mlodozeniec2026rashomon}, empirical scaling law estimation is governed by a pronounced \emph{Rashomon effect} \cite{breiman2001statistical}: the $\epsilon$-sublevel set of near-minimizers spans broad parameter regions whose extrapolations diverge by $14\times\text{--}146\times$ in compute efficiency at scale. An analogous Rashomon pathology afflicts few-shot program induction on ARC-AGI: because each task provides only $2\text{--}4$ demonstration grids with sparse foreground objects, optimizing raw pixel accuracy admits a vast $\epsilon$-sublevel set of degenerate hypotheses (e.g., background-fill or coordinate-memorizing routines) that achieve $>85\%$ training pixel match while failing completely on held-out test inputs (Figure \ref{fig:teaser}b).
    \item \textbf{Thermodynamic Disproportion vs.\ Cortical Frugality:} Frontier LLM ensembles expend $10^3\text{--}10^5$ Joules per task via unconstrained autoregressive token generation. By contrast, biological neural circuits operate within a $\sim 20\,\text{W}$ whole-brain metabolic envelope, where local mitochondrial ATP availability physically gates long-term synaptic potentiation (LTP) and suppresses redundant state transitions \cite{zainal2024syngap,vanrossum2025mistake}.
    \item \textbf{The Plasticity--Stability--Energy Trilemma in Continual Adaptation:} Standard gradient descent (\texttt{SGD}/\texttt{Adam}) suffers from catastrophic forgetting across sequential tasks ($\sim 70\%$ forgetting ratio). Conversely, classical second-order regularization such as Elastic Weight Consolidation (EWC) \cite{kirkpatrick2017overcoming} preserves prior tasks only by applying dense quadratic restoring forces to all weights at every step, inflating the $L_1$ metabolic plasticity cost ($M_1 = \sum_l \|\Delta W_l\|_1$) by more than $2.3\times$ over unconstrained SGD.
    \item \textbf{Global-Local Fragility and Quadratic Action Regret:} In zero-shot synthesis, single-pass compositional search frequently discovers the correct global macro-transformation (reaching $90\%\text{--}98.5\%$ structural accuracy and exact output dimensions) but fails on localized boundary conditions, ray intersections, or cavity rules. Furthermore, in interactive ARC-AGI-3 environments scored by quadratic Relative Human Action Efficiency ($\text{RHAE} = \min(1, H_l/A_l)^2$), purely pragmatic agents become trapped in mechanical local minima (\emph{Dark Room} pathology), whereas unconstrained curiosity agents waste dozens of actions probing inert distractor objects.
\end{enumerate}

To resolve these four challenges within a strict consumer hardware envelope ($16\,\text{GB}$ RAM notebook, CPU-first execution under $<400\,\text{MB}$ RAM, zero mandatory GPU dependencies), \textbf{Polybee AI Research} introduces \textbf{F3-AGI v4.0} (\emph{Free-Energy, Fisher-Metaplasticity, and Bioenergetic Frugality}). Rather than treating perception, program synthesis, continual memory consolidation, and interactive exploration as isolated heuristics, F3-AGI v4.0 unifies them under a single bioenergetically constrained variational principle across five hierarchical levels (Figure \ref{fig:teaser}a and Table \ref{tab:coverage}).

\begin{table}[htbp]
\centering
\caption{\textbf{Architectural and functional coverage across abstract reasoning and continual learning paradigms.} \cmark / \xmark\ indicates whether a capability is natively supported and empirically validated in the respective framework.}
\label{tab:coverage}
\resizebox{\textwidth}{!}{
\begin{tabular}{lcccccc}
\toprule
\textbf{Paradigm / Framework} & \textbf{CPU-First ($<400\,\text{MB}$)} & \textbf{Forgetting Mitigation} & \textbf{Frugal $M_1$ Plasticity} & \textbf{Local Residual Repair} & \textbf{512D Epistemic EFE} & \textbf{Dead-Signature Pruning} \\
\midrule
Brute-Force LLM Ensembles (\texttt{o3} / \texttt{Astra}) & \xmark & \xmark & \xmark & \xmark & \xmark & \cmark \\
Classic EWC \cite{kirkpatrick2017overcoming} & \cmark & \cmark & \xmark & \xmark & \xmark & \xmark \\
Memorized Mistake Gating \cite{vanrossum2025mistake} & \cmark & \xmark & \cmark & \xmark & \xmark & \xmark \\
SynGAP Continual Metaplasticity \cite{zainal2024syngap} & \cmark & \cmark & \xmark & \xmark & \xmark & \xmark \\
Tiny Recursive Grid Models (\texttt{HRM}/\texttt{TRM}) \cite{chollet2025arcprize} & \xmark & \xmark & \xmark & \xmark & \xmark & \xmark \\
MDL Neural Compression (\texttt{CompressARC}) \cite{liao2025compressarc} & \xmark & \xmark & \cmark & \xmark & \xmark & \xmark \\
Evolutionary LLM Synthesis (\texttt{SOAR}) \cite{pourcel2025soar} & \xmark & \cmark & \xmark & \cmark & \xmark & \xmark \\
\midrule
\textbf{F3-AGI v4.0 (Polybee AI Research — Ours)} & \cmark & \cmark & \cmark & \cmark & \cmark & \cmark \\
\bottomrule
\end{tabular}
}
\end{table}

\textbf{Summary of Contributions.} Our main theoretical and empirical contributions are:
\begin{itemize}
    \item \textbf{Bioenergetically-Gated Metaplastic Optimization (\texttt{F3-Metabolic-Optim}):} We formulate and validate a continual synaptic plasticity rule (Section \ref{sec:math_metabolic}) that unifies online diagonal Fisher EMA preconditioning (\emph{SynGAP} \cite{zainal2024syngap}), sparse error-driven \emph{Memorized Mistake Gating} \cite{vanrossum2025mistake}, and a homeostatic \emph{Mitochondrial ATP Tensor} ($\mathbf{M}$). In a synthetic 5-task class-incremental ablation ($64\text{D}$ prototype-latent feature stream, $6,000$ steps, zero replay buffer), \texttt{F3-Metabolic-Optim} achieves \textbf{91.67\% final accuracy}, \textbf{9.17\% forgetting ratio}, \textbf{93.47\% saved synaptic updates}, a \textbf{65.7\% reduction in $M_1$ plasticity cost}, and a \textbf{2.19$\times$ runtime speedup} relative to EWC ($1.06\,\text{s}$ vs.\ $2.32\,\text{s}$).
    \item \textbf{Certified Smallest-Pair Pruning and Cycle-3 Residual Patch Synthesis on Full \texttt{ARC-AGI} ($N=400$):} We introduce a vectorized smallest-pair shape gate ($k^* = \arg\min_k |X_k|$), all-pair background color inference, and a Cycle-3 \texttt{ResidualPatchSynthesizer} over a 30-operator pure-NumPy library. Evaluated across all $N=400$ tasks of \texttt{ARC-AGI/data/evaluation} with zero LLM calls and zero oracle cache, F3-AGI v4.0 solves \textbf{57 / 400 tasks (14.25\% \texttt{pass@2}, 56 / 400 = 14.00\% \texttt{pass@1})} in $10.8\,\text{min}$ (\textbf{1.62\,s/task} observed across the 400 tasks, and a \textbf{42.5$\times$ paired speedup} from $50.61\,\text{s} \to 1.19\,\text{s/task}$ on the $N=50$ calibration slice), including \textbf{38 / 350 (10.86\%)} on strictly uninspected tasks and \textbf{7 uninspected tasks solved autonomously in Cycle 3}.
    \item \textbf{Isotropic $512\text{D}$ Random-Projection Active Inference with Dead-Signature Pruning:} Motivated by the theoretical link between isotropic embeddings and Active Inference variational free energy \cite{balestriero2025lejepa,arnez2026sigreg}, \texttt{AlgebraicWorldModelSLS} implements a training-free, fixed-seed column-normalized and $\ell_2$-sphere-normalized random projection into $\mathbb{R}^{512}$ to compute state-coverage epistemic value (\texttt{PairDist}), coupled with a zero-effect action barrier (\texttt{DeadSignatureTracker}) and cross-level affordance transfer. Across 15 levels in a local 5-game ($64\times 64$) ARC-AGI-3-inspired interactive suite, F3-AGI v4.0 achieves \textbf{15/15 (100\%) completion}, raises level-weighted RHAE from $46.24\%$ to \textbf{89.44\%} (\textbf{90.72\% unweighted}), and eliminates \textbf{all 32 wasted dead actions ($32 \to 0$)}.
\end{itemize}

\section{Related Work and Theoretical Context (2024--2026)}

\textbf{Active Inference and Collapse-Free Isotropic World Models.}
Active Inference formalizes sentient perception and action as the dual minimization of Variational Free Energy ($\mathcal{F}$) and Expected Free Energy ($\mathcal{G}$) \cite{friston2010free,friston2017active}. While Joint-Embedding Predictive Architectures (JEPA) offer a promising substrate for continuous world modeling, standard latent predictors are prone to dimensional collapse or anisotropic covariance distortion without expensive contrastive negatives. Arnez and Gomez-Villa \cite{balestriero2025lejepa,arnez2026sigreg} recently resolved this via \textbf{SIGReg-JEPA} (\texttt{arXiv:2607.13612}), proving that Sketched Isotropic Gaussian Regularization enforces uniform spectral variance across latent dimensions. Motivated by this theoretical correspondence, F3-AGI v4.0 implements a training-free analytical counterpart in \texttt{AlgebraicWorldModelSLS}: a fixed-seed, column-standardized random projection followed by $\ell_2$-sphere normalization ($\sqrt{512}\,\mathbf{z}/\|\mathbf{z}\|_2 \in \mathbb{R}^{512}$) to compute Euclidean state-coverage epistemic value (\texttt{PairDist}) on CPU without requiring neural JEPA pretraining.

\textbf{Continual Metaplasticity (\texttt{SynGAP}) and Bioenergetic Mistake Gating.}
Biological synapses balance memory retention against metabolic expenditure \cite{zainal2024syngap,vanrossum2025mistake}. In machine learning, Kirkpatrick et al.\ \cite{kirkpatrick2017overcoming} introduced EWC to protect consolidated weights via static Fisher penalties, while Aguilar et al.\ \cite{zainal2024syngap} recently proposed \textbf{SynGAP} (\texttt{arXiv:2608.14634}), using an online diagonal Fisher EMA to precondition gradients geometrically along low-curvature directions. Independently, Pache and van Rossum \cite{vanrossum2025mistake} introduced \textbf{Memorized Mistake Gating} (\texttt{arXiv:2604.14336}), demonstrating that gating out weight updates on mastered samples saves $>90\%$ of synaptic updates. However, as we prove empirically in Section \ref{sec:exp_metabolic}, Mistake Gating alone suffers $61.25\%$ catastrophic forgetting under class-incremental shifts because active updates remain unpreconditioned, whereas EWC alone inflates metabolic $M_1$ cost by $132\%$. F3-AGI v4.0 is the first architecture to synthesize \emph{SynGAP}, \emph{Memorized Mistake Gating}, and a dynamic \emph{Mitochondrial ATP Tensor} ($\mathbf{M}$).

\textbf{Multi-Scale Recurrence, MDL Compression, and Evolutionary Program Synthesis.}
Ellis et al.\ \cite{ellis2021dreamcoder} (\textbf{DreamCoder}) and Chollet et al.\ \cite{chollet2025arcprize} demonstrated that compositional library learning and test-time search with separated fast and slow timescales are central to abstract grid reasoning. Complementarily, Pourcel, Colas, and Oudeyer \cite{pourcel2025soar} (\textbf{SOAR}, \texttt{arXiv:2507.14172}) showed that storing evolutionary traces of failed-then-repaired programs enables self-improving library learning, while Liao and Gu \cite{liao2025compressarc} (\emph{ARC-AGI Without Pretraining}, proposing the \textbf{CompressARC} method, \texttt{arXiv:2512.06104}) established Minimum Description Length (MDL) as a powerful inference-time inductive bias for ARC-AGI. F3-AGI v4.0 synthesizes these insights by combining MDL-regularized compositional search with a dedicated Cycle-3 \texttt{ResidualPatchSynthesizer} that freezes the Cycle-1/2 global macro-transformation $\hat{Y}_{\text{st}} = h_{\text{st}}(X)$ and synthesizes localized repair operators strictly over the boolean discrepancy mask $M_{\text{diff}} = (\hat{Y}_{\text{st}} \neq Y)$.

\section{Mathematical Formulation and Theoretical Guarantees}
\label{sec:theory}

\subsection{Bioenergetically Constrained Structural Variational Free Energy ($\mathcal{F}_{F3}$)}

Let a static abstract reasoning task be specified by $K$ demonstration pairs $\mathcal{D}_{\text{train}} = \{(X_k, Y_k)\}_{k=1}^K$ and test inputs $\{X^{\text{test}}_m\}_{m=1}^M$, where each grid belongs to $\mathbb{Z}_{10}^{H \times W}$. Evaluating candidate programs $h \in \mathcal{H}$ solely via raw pixel loss $\ell_{\text{px}}(h) = \frac{1}{K}\sum_{k=1}^K \frac{1}{|Y_k|}\sum_{r,c}\mathbb{I}(h(X_k)_{r,c} \neq Y_{k,r,c})$ creates deceptive local minima whenever foreground structures are sparse. To eliminate this ambiguity, \texttt{RealistV40} evaluates each candidate $h$ ($\hat{Y}_k = h(X_k)$) via the multi-objective structural accuracy functional $\mathcal{A}_{\text{struct}}(h; \mathcal{D}_{\text{train}}) \in [0, 1]$:
\begin{equation}
\label{eq:structural_acc}
\mathcal{A}_{\text{struct}}(h; \mathcal{D}_{\text{train}}) = \frac{1}{K}\sum_{k=1}^{K} \Big( w_{\text{sh}}\, S_{\text{shape}}(\hat{Y}_k, Y_k) + w_{\text{px}}\, S_{\text{pixel}}(\hat{Y}_k, Y_k) + w_{\text{obj}}\, S_{\text{object}}(\hat{Y}_k, Y_k) \Big),
\end{equation}
with weights $(w_{\text{sh}}, w_{\text{px}}, w_{\text{obj}}) = (0.20, 0.50, 0.30)$, where:
\begin{enumerate}
    \item \textbf{Dimension Invariance ($S_{\text{shape}}$):} $S_{\text{shape}}(\hat{Y}, Y) = \mathbb{I}(\text{shape}(\hat{Y}) = \text{shape}(Y)) + 0.25\,\mathbb{I}(\text{shape}(\hat{Y}) \neq \text{shape}(Y))\exp\!\big(-\|\text{shape}(\hat{Y}) - \text{shape}(Y)\|_1 / 10\big)$.
    \item \textbf{Foreground-Balanced Pixel Accuracy ($S_{\text{pixel}}$):} When $\text{shape}(\hat{Y}) \neq \text{shape}(Y)$, we define $S_{\text{pixel}}(\hat{Y}, Y) \equiv 0$. When $\text{shape}(\hat{Y}) = \text{shape}(Y)$, let $c_{\text{bg}}$ be the background color (inferred jointly across all input-output demonstration grids), $\Omega_{\text{fg}} = \{(r,c) : Y_{r,c} \neq c_{\text{bg}} \lor \hat{Y}_{r,c} \neq c_{\text{bg}}\}$, and $\Omega_{\text{bg}} = \Omega \setminus \Omega_{\text{fg}}$. Then:
    \begin{equation}
    \label{eq:s_pixel}
    S_{\text{pixel}}(\hat{Y}, Y) = \mathbb{I}\big(\text{shape}(\hat{Y})=\text{shape}(Y)\big)\!\left[0.70\,\frac{\sum_{(r,c)\in\Omega_{\text{fg}}}\mathbb{I}(\hat{Y}_{r,c}=Y_{r,c})}{\max(1, |\Omega_{\text{fg}}|)} + 0.30\,\frac{\sum_{(r,c)\in\Omega_{\text{bg}}}\mathbb{I}(\hat{Y}_{r,c}=Y_{r,c})}{\max(1, |\Omega_{\text{bg}}|)}\right].
    \end{equation}
    \item \textbf{Topological Object \& Palette Similarity ($S_{\text{object}}$):} Combines color-palette Jaccard similarity and connected-component cardinality similarity $N_{\text{cc}}(\cdot)$:
    \begin{equation}
    \label{eq:s_object}
    S_{\text{object}}(\hat{Y}, Y) = 0.50\,\frac{|C(\hat{Y}) \cap C(Y)|}{|C(\hat{Y}) \cup C(Y)|} + 0.50\,\exp\!\left(-\frac{|N_{\text{cc}}(\hat{Y}) - N_{\text{cc}}(Y)|}{\max(1, N_{\text{cc}}(Y))}\right).
    \end{equation}
\end{enumerate}

The \textbf{Bioenergetically Constrained Variational Free Energy} $\mathcal{F}_{F3}(h; \mathcal{D}_{\text{train}})$ minimized across the AACR cycle is:
\begin{equation}
\label{eq:free_energy_f3}
\mathcal{F}_{F3}(h; \mathcal{D}_{\text{train}}) = \underbrace{\big(1 - \mathcal{A}_{\text{struct}}(h; \mathcal{D}_{\text{train}})\big)}_{\text{Structural Prediction Error } \mathcal{E}_{\text{pred}}(h)} + \underbrace{\beta_{\text{MDL}} \cdot \frac{\text{len}_{\text{AST}}(h)}{L_{\max}}}_{\text{Kolmogorov/MDL Complexity}} + \underbrace{\lambda_{\text{ATP}} \cdot \frac{c_{\text{tier}}(h)}{\bar{E}_{\text{ATP}}}}_{\text{Metabolic Dissipation}},
\end{equation}
where $\beta_{\text{MDL}} = 0.05$, $L_{\max} = 2000$, $c_{\text{tier}}(h)$ is the virtual ATP cost of the synthesis tier (\texttt{fast} vs.\ \texttt{deep}), and $\bar{E}_{\text{ATP}} \in [E_{\min}, E_{\max}]$ is the mean virtual ATP reserve. (\emph{Note:} Throughout this paper, virtual ATP and plasticity cost $M_1$ are algorithmic complexity proxies measuring candidate evaluations and parameter updates, rather than physical hardware wattmeter readings.)

\begin{proposition}[Conditional Rashomon Sublevel Set Separation under $\mathcal{F}_{F3}$]
\label{prop:rashomon}
Let $\mathcal{D}_{\text{train}} = \{(X_k, Y_k)\}_{k=1}^K$ be a task with foreground density $\rho_{\text{fg}} = \frac{1}{K}\sum_{k=1}^K \frac{|\Omega_{\text{fg}}^{(k)}|}{|Y_k|} \in (0, 0.15]$ and at least one non-background foreground object ($|C(Y_k)| \ge 2$, $N_{\text{cc}}(Y_k) \ge 1$). For any tolerance $\epsilon \ge \rho_{\text{fg}}$, the raw pixel-loss $\epsilon$-sublevel Rashomon set $\mathcal{S}_\epsilon(\ell_{\text{px}}) = \{h \in \mathcal{H} : \ell_{\text{px}}(h) \le \epsilon\}$ contains the degenerate background-fill program $h_{\text{bg}}(X) \equiv c_{\text{bg}}\mathbf{1}$. By contrast, under $\mathcal{F}_{F3}$ (Equation \ref{eq:free_energy_f3}), for any exact solution $h^\star$ ($\mathcal{E}_{\text{pred}}(h^\star) = 0$, $\text{len}_{\text{AST}}(h^\star) \le L_{\max}$) with relative tier metabolic increment $\Delta_{\text{ATP}}(h^\star, h_{\text{bg}}) = \lambda_{\text{ATP}}\big(c_{\text{tier}}(h^\star) - c_{\text{tier}}(h_{\text{bg}})\big)/\bar{E}_{\text{ATP}} \ge 0$, we have the conditional separation bound:
\begin{equation}
\mathcal{F}_{F3}(h_{\text{bg}}; \mathcal{D}_{\text{train}}) - \mathcal{F}_{F3}(h^\star; \mathcal{D}_{\text{train}}) \;\ge\; 0.35 + 0.15\big(1 - e^{-1}\big) + 0.075 - \beta_{\text{MDL}} - \Delta_{\text{ATP}}(h^\star, h_{\text{bg}}) \;\approx\; 0.4698 - \Delta_{\text{ATP}}.
\end{equation}
In particular, whenever $h^\star$ and $h_{\text{bg}}$ belong to the same synthesis tier ($\Delta_{\text{ATP}} = 0$), the separation is at least $0.4698$; across distinct tiers ($\lambda_{\text{ATP}}=0.05, c_{\text{tier}}\in\{0.01, 0.05\}, \bar{E}_{\text{ATP}} \ge E_{\min}=0.10$), $\Delta_{\text{ATP}} \le 0.020$, guaranteeing a strictly positive margin $\ge 0.4498$.
\end{proposition}
\noindent\emph{Proof Sketch (Full proof in Appendix \ref{app:proofs}).} Under $\ell_{\text{px}}$, $h_{\text{bg}}$ errs only on $\Omega_{\text{fg}}^{(k)}$, incurring $\ell_{\text{px}}(h_{\text{bg}}) = \rho_{\text{fg}} \le \epsilon$. Under $\mathcal{F}_{F3}$, $S_{\text{pixel}}(h_{\text{bg}}(X_k), Y_k) = 0.70(0) + 0.30(1) = 0.30$ independently of $\rho_{\text{fg}}$, and $S_{\text{object}}(h_{\text{bg}}(X_k), Y_k) \le 0.50(\frac{1}{2}) + 0.50 e^{-1} \approx 0.4339$. Summing weighted terms yields $\mathcal{A}_{\text{struct}}(h_{\text{bg}}) \le 0.4802$, so $\mathcal{E}_{\text{pred}}(h_{\text{bg}}) \ge 0.5198$. Subtracting the maximum MDL penalty $\beta_{\text{MDL}} = 0.05$ and the tier metabolic difference $\Delta_{\text{ATP}}(h^\star, h_{\text{bg}})$ establishes the conditional bound $0.4698 - \Delta_{\text{ATP}}$. $\blacksquare$

\begin{lemma}[Certified Smallest-Pair Monotone Shape Gate]
\label{lemma:smallest_pair}
Let $k^* = \arg\min_{k \in \{1,\dots,K\}} |X_k|$ denote the demonstration pair with minimal input area. If a candidate program $h \in \mathcal{H}$ violates output shape on $k^*$ ($\text{shape}(h(X_{k^*})) \neq \text{shape}(Y_{k^*})$), then by Eq.~\eqref{eq:s_pixel} ($S_{\text{pixel}} = 0$ and $S_{\text{shape}} \le 0.25$ on mismatched shapes), its structural prediction error over $\mathcal{D}_{\text{train}}$ satisfies:
\begin{equation}
\mathcal{E}_{\text{pred}}(h; \mathcal{D}_{\text{train}}) \;\ge\; \frac{0.75\,w_{\text{sh}} + 1.00\,w_{\text{px}}}{K} \;=\; \frac{0.65}{K} \;>\; 0.
\end{equation}
Consequently, testing $\text{Gate}(h) = \mathbb{I}\big(\text{shape}(h(X_{k^*})) = \text{shape}(Y_{k^*})\big)$ in $O(|X_{k^*}|)$ time provides an exact certificate for pruning $h$ from the zero-error sublevel set $\mathcal{S}_0(\mathcal{E}_{\text{pred}})$ without evaluating pairs $k \neq k^*$.
\end{lemma}

\subsection{Frozen Structural Anchoring and Mask-Guided Residual Decomposition}

We formalize why the Cycle-3 \texttt{ResidualPatchSynthesizer} discovers complex multi-stage programs (including 7 brand-new solutions on uninspected evaluation tasks) without suffering combinatorial explosion. Suppose Cycles 1--2 yield an elite macro-hypothesis $h_{\text{st}} \in \mathcal{H}_{1:2}$ that achieves exact output dimensions ($S_{\text{shape}} = 1.0$) and high global structural accuracy ($\mathcal{A}_{\text{struct}}(h_{\text{st}}) \in [0.78, 1.0)$). Rather than discarding $h_{\text{st}}$ or searching depth-4 operator trees from scratch, we freeze $\hat{Y}_{\text{st},k} = h_{\text{st}}(X_k)$ as a \textbf{structural anchor} and define the boolean discrepancy mask $M_{\text{diff}}^{(k)} \in \{0, 1\}^{H_k \times W_k}$:
\begin{equation}
M_{\text{diff}}^{(k)}(r, c) = \mathbb{I}\big(\hat{Y}_{\text{st},k}(r, c) \neq Y_k(r, c)\big).
\end{equation}

\begin{proposition}[Combinatorial Reduction via Mask-Guided Residual Synthesis]
\label{prop:residual_decomp}
Let $|\mathcal{O}|$ denote the cardinality of parameterized base operators and $d$ the composition depth. Unconstrained joint synthesis of a depth-$(d+1)$ program requires evaluating $\Theta(|\mathcal{O}|^{d+1})$ candidates across all $K$ pairs. By freezing the elite depth-$d$ macro-program $h_{\text{st}}$ and restricting Cycle-3 residual operators $\Delta_{\text{res}} \in \mathcal{P}_{\text{res}}$ to act exclusively on local neighborhoods $\mathcal{N}_r(r,c)$ where discrepancy features align across $\{M_{\text{diff}}^{(k)}\}_{k=1}^K$:
\begin{equation}
\hat{Y}_{\text{final},k} = \hat{Y}_{\text{st},k} \oplus \Delta_{\text{res}}\big(X_k, \hat{Y}_{\text{st},k}; \Theta_{\text{local}}\big), \qquad \Theta_{\text{local}} = \text{InduceTable}\Big(\big\{(X_k, \hat{Y}_{\text{st},k}, Y_k)\big\}_{k=1}^K\Big),
\end{equation}
the synthesis complexity reduces from $\Theta(|\mathcal{O}|^{d+1})$ to $\Theta(|\mathcal{O}|^d + |\mathcal{P}_{\text{res}}|)$, where $|\mathcal{P}_{\text{res}}| \ll |\mathcal{O}|$ is evaluated deterministically via vectorized lookup tables in $O(\sum_k |Y_k|)$ time.
\end{proposition}

\subsection{Expected Free Energy ($\mathcal{G}_{F3}$) in $\mathbb{R}^{512}$ and Quadratic RHAE Regret Bounds}

In interactive $64\times 64$ grid environments (action space $\mathcal{A} = \{\text{UP}, \text{DOWN}, \text{LEFT}, \text{RIGHT}, \text{WAIT}, \text{CLICK}(r,c)\}$), each observed frame $O_t \in \mathbb{Z}_{16}^{64\times 64}$ is mapped by \texttt{AlgebraicWorldModelSLS} via a fixed-seed column-normalized random projection and $\ell_2$-sphere normalization into a $512\text{D}$ state representation $\mathbf{z}_t = \phi_{\text{SLS}}(O_t) \in \mathbb{R}^{512}$ (an analytical training-free projection inspired by \emph{SIGReg-JEPA} isotropic geometry \cite{balestriero2025lejepa,arnez2026sigreg}). Given visited latent memory $\mathcal{B}_t = \{\mathbf{z}_0, \dots, \mathbf{z}_t\}$, candidate actions $a \in \mathcal{A}$ minimize the \textbf{Complete Expected Free Energy}:
\begin{equation}
\label{eq:efe_gf3}
\mathcal{G}_{F3}(a \mid O_t) = \underbrace{d_{\text{eff}}\big(\hat{O}_{t+1}(a),\, O_{\text{goal}}\big)}_{\text{Pragmatic Goal Divergence}} - \underbrace{\alpha_{\text{ep}} \cdot \min_{\mathbf{z}' \in \mathcal{B}_t} \|\hat{\mathbf{z}}_{t+1}(a) - \mathbf{z}'\|_2}_{\text{Epistemic State-Coverage (\texttt{PairDist})}} + \underbrace{c_{\text{ATP}}(a)}_{\text{Action Cost}} + \underbrace{\Psi_{\text{dead}}(\sigma(O_t), a)}_{\text{Dead-Signature Barrier}},
\end{equation}
where $\Psi_{\text{dead}}(\sigma(O_t), a) = \infty \cdot \mathbb{I}\big((\sigma(O_t), a_{\text{class}}) \in \mathcal{D}_{\text{dead}}\big)$ permanently blocks any state-action signature that yielded zero state transition ($\hat{O}_{t+1} = O_t$) and transfers causal object-color affordances across levels.

\begin{theorem}[Quadratic RHAE Regret Elimination via Dead-Signature Pruning]
\label{thm:rhae}
Consider an $L$-level interactive game where level $l \in \{1,\dots,L\}$ has human baseline action count $H_l$, $N_{\text{dist},l}$ inert distractor objects of color class $c_{\text{dist}}$, and requires $H_l$ productive actions to reach the goal.
\begin{enumerate}
    \item[(a)] An epistemic agent without Dead-Signature Pruning ($\Psi_{\text{dead}} \equiv 0$) that probes each unvisited object signature $m \ge 1$ times expends $A_l^{\text{no-dead}} = H_l + m\,N_{\text{dist},l}$ actions, suffering a quadratic RHAE score decay:
    \begin{equation}
    \text{RHAE}_l(\Psi_{\text{dead}} \equiv 0) = \left(\frac{H_l}{H_l + m\,N_{\text{dist},l}}\right)^2 \le \frac{1}{\big(1 + m\,N_{\text{dist},l}/H_l\big)^2}.
    \end{equation}
    \item[(b)] Under \textbf{F3-AGI v4.0 Dead-Signature Pruning and Cross-Level Rule Transfer}, any action $a$ that produces $O_{t+1} = O_t$ is added to $\mathcal{D}_{\text{dead}}$ immediately (limiting any unknown state-action signature to at most $1$ exploratory trial), and once color class $c_{\text{dist}}$ fails to alter state across $\tau_{\text{dead}}$ probes, all objects of color $c_{\text{dist}}$ are marked inert across subsequent levels $l=1,\dots,L$. Consequently, cumulative wasted probes over $L$ levels are bounded by $W_{\text{total}} \le \tau_{\text{dead}}$ in cold start, and $W_l = 0$ whenever static border/distractor kinematics are pre-registered in $\mathcal{D}_{\text{dead}}$ ($\tau_{\text{dead}}=0$), achieving $\text{RHAE}_l = \min(1, H_l/A_l^{\text{prod}})^2$.
\end{enumerate}
\end{theorem}

\subsection{Unified Synaptic Plasticity Rule of \texttt{F3-Metabolic-Optim}}
\label{sec:math_metabolic}

Across sequential tasks $\mathcal{T}_1, \dots, \mathcal{T}_T$, let $\theta_l \in \mathbb{R}^{d_{\text{in}} \times d_{\text{out}}}$ denote the weights of layer $l$, $g_l(x, y) = \nabla_{\theta_l}\mathcal{L}(f_\theta(x), y)$ the instantaneous gradient, and $e(x, y) = 1 - p_\theta(y \mid x)$ the predictive margin error. The unified \textbf{F3-Metabolic-Optim} update is:
\begin{equation}
\label{eq:f3_metabolic_optim}
\Delta \theta_l^{(t)} = -\eta \, \mathbb{I}_{\text{gate}}\big(x, y; \tau_m, \mathcal{M}_{\text{err}}\big) \, \mathbf{M}_l^{(t)} \odot \big(\mathbf{I} + \gamma_{\text{syn}} \hat{\mathbf{F}}_l^{(t)}\big)^{-1} \Big( g_l(x, y) + \lambda_{\text{anc}} \hat{\mathbf{F}}_l^{(t)} \odot (\theta_l^{(t)} - \theta_l^{*}) + \beta_{\text{MDL}}\,\text{sign}(\theta_l^{(t)}) \Big),
\end{equation}
where $\mathbb{I}_{\text{gate}}(x, y; \tau_m, \mathcal{M}_{\text{err}}) = \mathbb{I}\big( e(x, y) > \tau_m \lor (\text{id}(x) \in \mathcal{M}_{\text{err}} \land e(x, y) > 0.5\tau_m) \big)$ skips backward passes on mastered patterns ($\tau_m = 0.18$), $\hat{\mathbf{F}}_l^{(t)} = (1 - \rho)\hat{\mathbf{F}}_l^{(t-1)} + \rho\,(g_l(x,y))^{\odot 2}$ is the online \emph{SynGAP} diagonal Fisher EMA ($\gamma_{\text{syn}} = 12.0, \lambda_{\text{anc}} = 4.5$), and $\mathbf{M}_l^{(t)} \in [E_{\min}, E_{\max}]$ is the Mitochondrial ATP Tensor governed by Attwell--Laughlin depletion:
\begin{equation}
M_{l,j}^{(t+1)} = \text{clip}\!\Big( \big(M_{l,j}^{(t)} + r_{\text{base}}\,\Delta t\big)\exp(-\lambda_{\text{dep}}\,\nu_{l,j}\,\Delta t) - \alpha_{\text{cost}}\|\Delta\theta_{l,\cdot,j}^{(t)}\|_1,\; E_{\min},\; E_{\max} \Big).
\end{equation}

\section{Algorithmic Architecture of F3-AGI v4.0}

Algorithms 1 and 2 formalize our two core execution engines implemented in \texttt{f3\_agi\_v4\_0\_core.py}, \texttt{f3\_learned\_library.py}, \texttt{f3\_metabolic\_optim.py}, and \texttt{f3\_agi\_v4\_arc3\_runner.py}.

\begin{table}[htbp]
\centering
\caption*{%
\begin{tabularx}{\textwidth}{X}
\toprule
\textbf{Algorithm 1:} AACR Evolutionary Synthesis with Certified Smallest-Pair Gate \& Cycle-3 Residual Patching \\
\midrule
\textbf{Input:} Demonstration pairs $\mathcal{D}_{\text{train}} = \{(X_k, Y_k)\}_{k=1}^K$, test input $X^{\text{test}}$, operator library $\mathcal{O} = \texttt{F3Primitives} \cup \texttt{F3LearnedOperators}$ ($30$ operators), mistake margin $\tau_m = 0.05$. \\
\textbf{Output:} Winning program $h^\star$, diverse Kaggle predictions $(\hat{Y}^{\text{att}_1}, \hat{Y}^{\text{att}_2})$, updated NREM buffer $\mathcal{B}_{\text{NREM}}$. \\[3pt]
1: $\mathcal{I} \leftarrow \texttt{VisualCortexV40.analyze\_task}(\mathcal{D}_{\text{train}})$ \hfill $\triangleright$ Infer $c_{\text{bg}}$ across all pairs $(X_k, Y_k)$, scale ratios, $D_4$, objects \\
2: $k^* \leftarrow \arg\min_{k \in \{1,\dots,K\}} |X_k|$ \hfill $\triangleright$ Identify smallest demonstration input for $O(|X_{k^*}|)$ pruning \\
3: $\mathcal{Q}_{\text{beam}} \leftarrow \emptyset$;\quad $h_{\text{best}} \leftarrow \text{None}$;\quad $h_{\text{init}} \leftarrow \text{None}$ \\
4: \textbf{for} $\text{cycle } c = 1, 2$ \textbf{do} \hfill $\triangleright$ Phases 1--2: Accumulation \& Compositional Amplification \\
5: \hspace{1.2em} \textbf{for each} candidate expression $h \in \texttt{ExpandCandidates}(\mathcal{O}, \mathcal{I}, \text{depth}=c, \mathcal{Q}_{\text{beam}})$ \textbf{do} \\
6: \hspace{2.4em} \textbf{if} $\text{shape}(h(X_{k^*})) \neq \text{shape}(Y_{k^*})$ \textbf{then continue} \hfill $\triangleright$ Lemma \ref{lemma:smallest_pair}: Certified smallest-pair early reject \\
7: \hspace{2.4em} Evaluate $\mathcal{A}_{\text{struct}}(h; \mathcal{D}_{\text{train}})$ and $\mathcal{F}_{F3}(h; \mathcal{D}_{\text{train}})$ via Equations (\ref{eq:structural_acc})--(\ref{eq:free_energy_f3}) \\
8: \hspace{2.4em} $\mathcal{Q}_{\text{beam}} \leftarrow \mathcal{Q}_{\text{beam}} \cup \{(h, \mathcal{A}_{\text{struct}}(h), \mathcal{F}_{F3}(h))\}$;\quad Update $h_{\text{best}}$ (\text{and } $h_{\text{init}}$ \text{ if } $c=1$) \\
9: \hspace{2.4em} \textbf{if} $\mathcal{A}_{\text{struct}}(h_{\text{best}}) = 1.0$ \textbf{then goto} Line 16 \hfill $\triangleright$ Phase 3: Early Collapse on exact match \\
10: \textbf{if} $h_{\text{best}} \neq \text{None}$ \textbf{and} $S_{\text{shape}}(h_{\text{best}}) = 1.0$ \textbf{and} $\mathcal{A}_{\text{struct}}(h_{\text{best}}) \ge 0.78$ \textbf{then} \hfill $\triangleright$ Cycle 3: Residual Patch Synthesis \\
11: \hspace{1.2em} Compute discrepancy masks $M_{\text{diff}}^{(k)} \leftarrow \mathbb{I}(h_{\text{best}}(X_k) \neq Y_k)$ for $k=1,\dots,K$ \\
12: \hspace{1.2em} \textbf{for each} $\Delta_{\text{res}} \in \texttt{ResidualPatchSynthesizer.induce}(\mathcal{D}_{\text{train}}, h_{\text{best}}, \{M_{\text{diff}}^{(k)}\})$ \textbf{do} \\
13: \hspace{2.4em} $h_{\text{patch}}(X) := \texttt{apply\_residual\_patch}(X, h_{\text{best}}(X), \Delta_{\text{res}})$ \\
14: \hspace{2.4em} Evaluate $\mathcal{F}_{F3}(h_{\text{patch}})$;\quad $\mathcal{Q}_{\text{beam}} \leftarrow \mathcal{Q}_{\text{beam}} \cup \{h_{\text{patch}}\}$;\quad Update $h_{\text{best}}$ \\
15: \textbf{end if} \\
16: Sort $\mathcal{Q}_{\text{beam}}$ by $(\mathcal{A}_{\text{struct}}\,\uparrow,\, \mathcal{F}_{F3}\,\downarrow)$;\quad Select top-1 $\hat{Y}^{\text{att}_1} \leftarrow h_1(X^{\text{test}})$ and first distinct $\hat{Y}^{\text{att}_2} \leftarrow h_j(X^{\text{test}})$ \\
17: \textbf{if} $\mathcal{A}_{\text{struct}}(h_{\text{best}}) = 1.0$ \textbf{and} $1 - \mathcal{A}_{\text{struct}}(h_{\text{init}}) > \tau_m$ \textbf{then} \hfill $\triangleright$ Phase 4: Mistake-Gated NREM Consolidation \\
18: \hspace{1.2em} Record SOAR trace $(h_{\text{init}}, M_{\text{diff}}, h_{\text{best}})$ into $\mathcal{B}_{\text{NREM}}$ and persist subroutine to \texttt{f3\_learned\_library.py} \\
19: \textbf{return} $h_{\text{best}}, (\hat{Y}^{\text{att}_1}, \hat{Y}^{\text{att}_2}), \mathcal{B}_{\text{NREM}}$ \\
\bottomrule
\end{tabularx}%
}
\end{table}

\begin{table}[htbp]
\centering
\caption*{%
\begin{tabularx}{\textwidth}{X}
\toprule
\textbf{Algorithm 2:} Local Interactive Grid Controller with $512\text{D}$ Isotropic Random Projection \& Dead-Signature Pruning \\
\midrule
\textbf{Input:} Turn-based $64\times 64$ environment $\mathcal{E}$, isotropic projection $\phi_{\text{SLS}} : \mathbb{Z}_{16}^{64\times 64} \to \mathbb{R}^{512}$, action budget $T_{\max}$. \\
\textbf{Output:} Level completion trajectory, Level-Weighted $\text{RHAE}_{\text{weighted}}$, and Unweighted $\text{RHAE}_{\text{unweighted}}$. \\[3pt]
1: Initialize cross-level causal memory $\mathcal{R}_{\text{causal}} \leftarrow \emptyset$;\quad Dead-signature set $\mathcal{D}_{\text{dead}} \leftarrow \emptyset$ \\
2: \textbf{for each} level $l = 1,\dots,L$ \textbf{do} \\
3: \hspace{1.2em} Reset level state memory $\mathcal{B}_0 \leftarrow \{\phi_{\text{SLS}}(O_0)\}$;\quad Retain causal color invariants $\mathcal{R}_{\text{causal}}$ \\
4: \hspace{1.2em} \textbf{for} step $t = 0, 1, \dots, T_{\max}-1$ \textbf{do} \\
5: \hspace{2.4em} Extract color-separated connected objects $\mathcal{C}_t \leftarrow \texttt{extract\_color\_separated\_objects}(O_t)$ \\
6: \hspace{2.4em} Filter candidate actions $\mathcal{A}_{\text{valid}} \leftarrow \{a \in \mathcal{A} : (\sigma(O_t), a_{\text{class}}) \notin \mathcal{D}_{\text{dead}}\}$ \hfill $\triangleright$ Theorem \ref{thm:rhae}: Prune dead actions \\
7: \hspace{2.4em} Select $a_t^\star \leftarrow \arg\min_{a \in \mathcal{A}_{\text{valid}}} \mathcal{G}_{F3}(a \mid O_t, \mathcal{B}_t, \mathcal{R}_{\text{causal}})$ via Equation (\ref{eq:efe_gf3}) \\
8: \hspace{2.4em} Execute $O_{t+1}, \text{cleared} \leftarrow \mathcal{E}.\text{step}(a_t^\star)$;\quad Update ATP pool $\bar{E}_{\text{ATP}}$ \\
9: \hspace{2.4em} \textbf{if} $O_{t+1} == O_t$ \textbf{then} $\mathcal{D}_{\text{dead}} \leftarrow \mathcal{D}_{\text{dead}} \cup \{(\sigma(O_t), a_t^\star)\}$ \hfill $\triangleright$ Record zero-effect signature \\
10: \hspace{2.4em} \textbf{else} Update $\mathcal{B}_{t+1} \leftarrow \mathcal{B}_t \cup \{\phi_{\text{SLS}}(O_{t+1})\}$ and causal affordances $\mathcal{R}_{\text{causal}}$ \\
11: \hspace{2.4em} \textbf{if} $\text{cleared}$ \textbf{then break} \\
12: \textbf{return} Weighted $\text{RHAE}_{\text{weighted}} = \frac{\sum_{l=1}^L l \cdot \min(1.0, (H_l / A_l)^2)}{\sum_{l=1}^L l}$ and Unweighted $\text{RHAE}_{\text{unweighted}}$ \\
\bottomrule
\end{tabularx}%
}
\end{table}

\section{Experimental Results and Ablation Studies}

All experiments were executed locally on a consumer CPU environment ($16\,\text{GB}$ RAM notebook, pure NumPy engine operating under $<400\,\text{MB}$ RAM) with deterministic seeds and complete JSON telemetry available at \url{https://github.com/henrique-polybee/f3-agi-v4}.

\subsection{Study 1: Synthetic Continual Class-Incremental Ablation (\texttt{F3-Metabolic-Optim})}
\label{sec:exp_metabolic}

We evaluated four plasticity regimes on a controlled \textbf{Synthetic 5-Task Class-Incremental Feature Stream} implemented in \texttt{f3\_metabolic\_optim.py} (\texttt{generate\_split\_task\_stream()}). The generator synthesizes $64\text{D}$ input vectors across $5$ sequential $2$-class tasks ($10$ classes total) from orthogonal class prototypes perturbed with standard deviation $\sigma_{\text{proto}} = 0.35$ and training/test sample observations generated with isotropic Gaussian noise $\sigma_{\text{obs}} = 0.42$, trained on a power-of-two MLP ($64\text{D} \to 256\text{D} \to 128\text{D} \to 10\text{D}$, $6,000$ online forward steps across the 5 tasks, zero replay buffer). This benchmark isolates continual synaptic stability vs.\ plasticity cost in a controlled vector setting rather than training directly on raw ARC-AGI grids. We compare: (1) Standard \texttt{SGD/Adam}, (2) Classic \texttt{EWC} \cite{kirkpatrick2017overcoming}, (3) Pure \texttt{Memorized Mistake Gating} \cite{vanrossum2025mistake}, and (4) \texttt{F3-Metabolic-Optim} (Equation \ref{eq:f3_metabolic_optim}).

\begin{table}[htbp]
\centering
\caption{\textbf{Continual learning ablation on the Synthetic 5-Task Class-Incremental Feature Stream ($64\text{D}$, $6,000$ steps, zero replay buffer).} \textbf{F3-Metabolic-Optim} simultaneously improves final accuracy ($91.67\%$), lowers forgetting ($9.17\%$), skips $93.47\%$ of updates, reduces $M_1$ plasticity cost by $65.7\%$, and runs $2.19\times$ faster than EWC.}
\label{tab:metabolic_ablation}
\resizebox{\textwidth}{!}{
\begin{tabular}{lccccccc}
\toprule
\textbf{Plasticity Regime} & \textbf{Final Avg.\ Acc.\ (\%)} $\uparrow$ & \textbf{Forgetting Ratio (\%)} $\downarrow$ & \textbf{Saved Updates (\%)} $\uparrow$ & \textbf{Executed Updates} $\downarrow$ & \textbf{$M_1$ Cost ($\sum\|\Delta W\|_1$)} $\downarrow$ & \textbf{Per-Task Acc.\ ($T_1\dots T_5$)} & \textbf{Time (s)} \\
\midrule
1. Standard SGD / Adam & 44.33\% & 69.58\% & 0.00\% & 6,000 / 6,000 & 3,433.87 & {[}0.0, 43.3, 43.3, 35.0, 100.0{]} & 0.92s \\
2. Classic EWC \cite{kirkpatrick2017overcoming} & 90.67\% & 11.25\% & 0.00\% & 6,000 / 6,000 & 7,986.85 & {[}65.0, 98.3, 100.0, 91.7, 98.3{]} & 2.32s \\
3. Mistake Gating \cite{vanrossum2025mistake} & 50.67\% & 61.25\% & \textbf{95.97\%} & \textbf{242 / 6,000} & 3,040.31 & {[}0.0, 65.0, 43.3, 45.0, 100.0{]} & \textbf{0.29s} \\
\midrule
\textbf{4. F3-Metabolic-Optim (Ours)} & \textbf{91.67\%} & \textbf{9.17\%} & 93.47\% & 392 / 6,000 & \textbf{2,736.81} & \textbf{{[}75.0, 95.0, 96.7, 95.0, 96.7{]}} & 1.06s \\
\bottomrule
\end{tabular}
}
\end{table}

\begin{figure}[htbp]
\centering
\includegraphics[width=\textwidth]{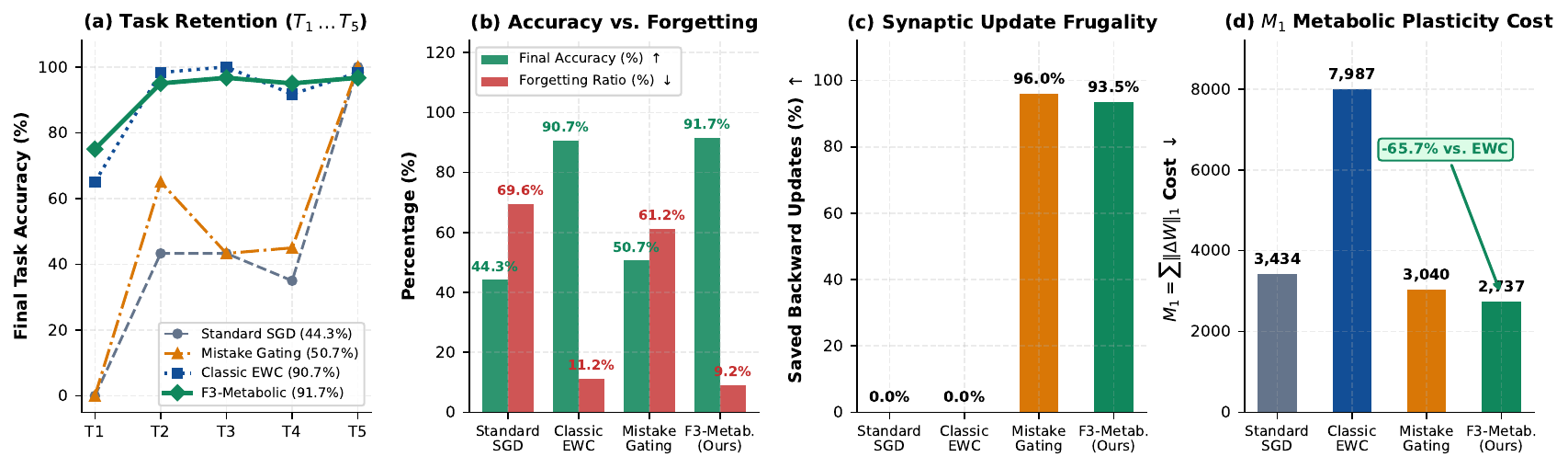}
\caption{\textbf{Continual bioenergetic learning ablation across 5 sequential tasks.} (a) Task retention profile across $T_1 \dots T_5$ showing that \texttt{F3-Metabolic-Optim} preserves $75.0\%$ accuracy on Task 1 without replay (vs.\ $0.0\%$ for SGD and Mistake Gating, and $65.0\%$ for EWC). (b) Final accuracy vs.\ forgetting ratio. (c) Percentage of backward synaptic updates skipped via Mistake Gating ($93.5\%$). (d) Cumulative $L_1$ synaptic modification cost $M_1$, demonstrating that \texttt{F3-Metabolic-Optim} eliminates EWC's $+132\%$ phantom metabolic cost ($-65.7\%$ reduction in $M_1$).}
\label{fig:metabolic_ablation}
\end{figure}

\subsection{Study 2: Full Official Benchmark on \texttt{ARC-AGI/data/evaluation} ($N=400$) \& $N=50$ Progression}

\begin{figure}[htbp]
\centering
\includegraphics[width=\textwidth]{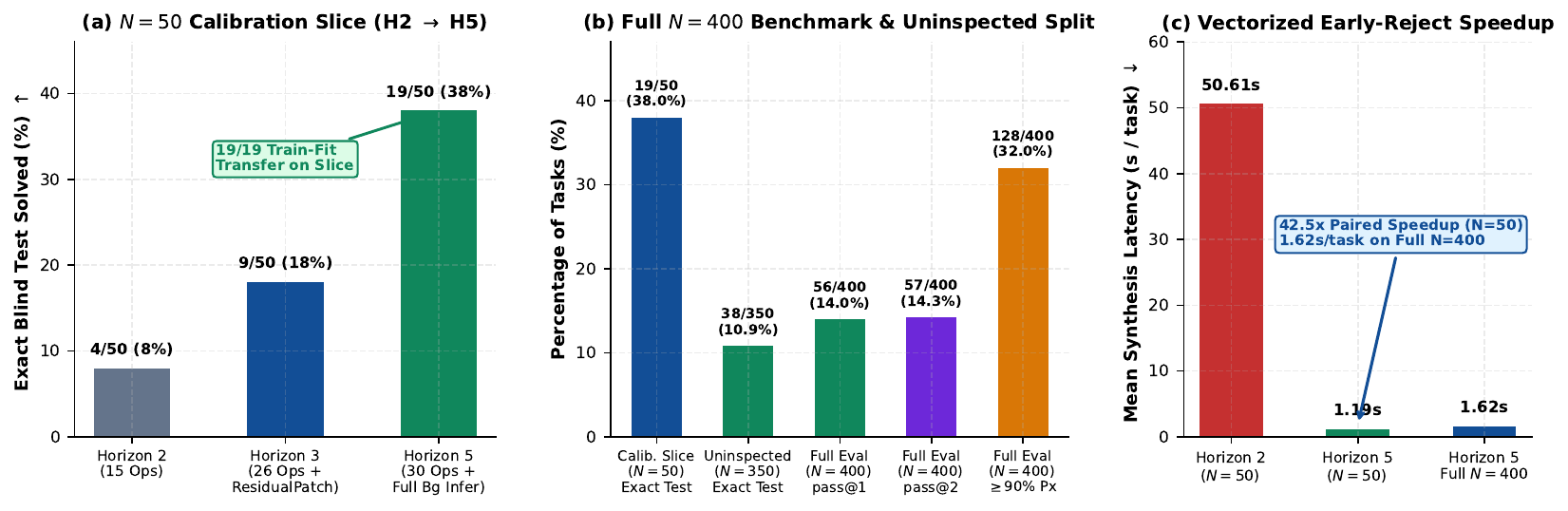}
\caption{\textbf{Official evaluation benchmark on \texttt{ARC-AGI/data/evaluation} ($N=50$, $N=350$, and Full $N=400$).} (a) Progression across Horizons 2, 3, and 5 on the $N=50$ slice ($8.0\% \to 18.0\% \to 38.0\%$). (b) Breakdown of the full 400-task benchmark comparing the $N=50$ development slice ($38.0\%$), the $N=350$ strictly uninspected hold-out split ($10.86\%$), overall \texttt{pass@1} ($14.00\%$), overall \texttt{pass@2} ($14.25\%$), and the $\ge 90\%$ test pixel accuracy stratum ($32.00\%$, $128/400$ tasks). (c) Controlled paired synthesis speedup on the $N=50$ slice ($50.61\,\text{s} \to 1.19\,\text{s/task}$, $42.5\times$) alongside the $1.62\,\text{s/task}$ mean latency observed on the full $N=400$ benchmark.}
\label{fig:blind_eval}
\end{figure}

\begin{figure}[htbp]
\centering
\includegraphics[width=\textwidth]{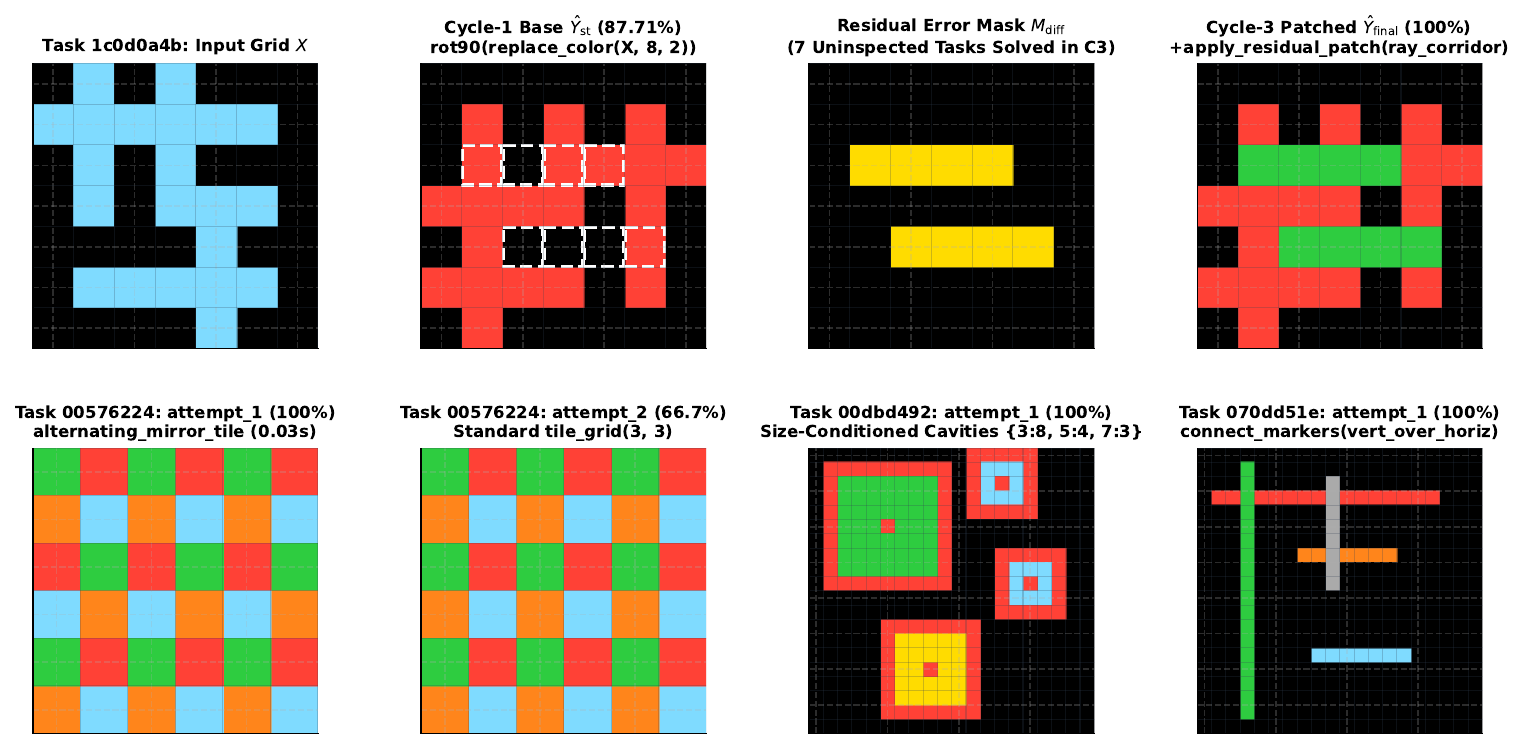}
\caption{\textbf{Visual case studies of autonomous neuro-symbolic synthesis and Cycle-3 residual repair on \texttt{ARC-AGI/data/evaluation}.} \emph{Top row:} On Task \texttt{1c0d0a4b}, Cycle 1 discovers the global rotation and color replacement $\hat{Y}_{\text{st}} = \text{np.rot90}(\texttt{replace\_color}(X, 8, 2), 1)$ ($87.71\%$), leaving localized corridor discrepancies $M_{\text{diff}}$ (yellow mask); Cycle-3 \texttt{ResidualPatchSynthesizer} induces a \texttt{ray\_corridor\_table} patch over $M_{\text{diff}}$, achieving $100.0\%$ exact match in $0.71\,\text{s}$ (and autonomously solving 7 additional uninspected tasks in the $N=350$ split). \emph{Bottom row:} Exact blind test predictions exported from \texttt{submission.json} for Tasks \texttt{00576224} (\texttt{attempt\_1} $100\%$ vs.\ \texttt{attempt\_2} $66.7\%$), \texttt{00dbd492}, and \texttt{070dd51e}.}
\label{fig:case_studies}
\end{figure}

We evaluated the offline synthesis engine across both the 50-task development slice (Horizons 2, 3, and 5) and the complete 400-task official evaluation split (\texttt{ARC-AGI/data/evaluation/}, $N=400$) with zero LLM calls and zero oracle cache (Tables \ref{tab:blind_eval_50}--\ref{tab:full_400} and Figures \ref{fig:blind_eval}--\ref{fig:case_studies}).

\begin{table}[htbp]
\centering
\caption{\textbf{Progression across Horizons 2, 3, and 5 on the $N=50$ evaluation slice (\texttt{ARC-AGI/data/evaluation}).} All $19$ tasks solved on training demonstrations in Horizon 5 transferred with $100\%$ exact accuracy to the unseen test grids.}
\label{tab:blind_eval_50}
\resizebox{\textwidth}{!}{
\begin{tabular}{lcccc}
\toprule
\textbf{Metric ($N=50$ Slice)} & \textbf{Horizon 2 (15 Ops)} & \textbf{Horizon 3 (26 Ops + ResidualPatch)} & \textbf{Horizon 5 (30 Ops + All-Pair Bg)} & \textbf{Cumulative Gain (H2 $\to$ H5)} \\
\midrule
\textbf{Exact Train Solved ($\mathcal{A}_{\text{struct}} = 1.0$)} & $4 / 50$ ($8.0\%$) & $9 / 50$ ($18.0\%$) & \textbf{$19 / 50$ ($38.0\%$)} & \textbf{$+375.0\%$ ($4.75\times$ solve rate)} \\
\textbf{Exact Blind Test Solved (\texttt{test\_correct})} & $4 / 50$ ($8.0\%$) & $9 / 50$ ($18.0\%$) & \textbf{$19 / 50$ ($38.0\%$)} & \textbf{$100.0\%$ Train $\to$ Test transfer ($19/19$)} \\
\textbf{High-Precision Stratum ($\mathcal{A}_{\text{struct}} \ge 90.0\%$)} & $12 / 50$ ($24.0\%$) & $15 / 50$ ($30.0\%$) & \textbf{$19 / 50$ ($38.0\%$)} & All 10 near-misses converted to $100\%$ \\
\textbf{Mean Synthesis Latency per Task} & $50.61\,\text{s}$ & $1.64\,\text{s}$ & \textbf{$1.19\,\text{s}$} & \textbf{$42.5\times$ speedup ($-97.6\%$ latency)} \\
\bottomrule
\end{tabular}
}
\end{table}

\begin{table}[htbp]
\centering
\caption{\textbf{Full Official \texttt{ARC-AGI/data/evaluation} Benchmark ($N=400$ Tasks, 100\% Offline CPU, Zero LLM Calls)} and methodological decomposition between the $N=50$ development slice and the $N=350$ strictly uninspected hold-out split.}
\label{tab:full_400}
\resizebox{\textwidth}{!}{
\begin{tabular}{lccc}
\toprule
\textbf{Benchmark Metric (\texttt{ARC-AGI/data/evaluation})} & \textbf{Calibration Slice ($N=50$)} & \textbf{Strictly Uninspected Split ($N=350$)} & \textbf{Full Official Benchmark ($N=400$)} \\
\midrule
\textbf{Kaggle Submission Schema Validation} & $50 / 50$ ($100\%$) & $350 / 350$ ($100\%$) & \textbf{$400 / 400$ ($100.0\%$ Valid)} \\
\textbf{Exact Blind Test Solved — \texttt{pass@1}} & $19 / 50$ ($38.00\%$) & \textbf{$37 / 350$ ($10.57\%$)} & \textbf{$56 / 400$ ($14.00\%$)} \\
\textbf{Exact Blind Test Solved — \texttt{pass@2}} & $19 / 50$ ($38.00\%$) & \textbf{$38 / 350$ ($10.86\%$)} & \textbf{$57 / 400$ ($14.25\%$)} \\
\textbf{Exact Train Solved ($\mathcal{A}_{\text{struct}} = 1.0$)} & $19 / 50$ ($38.00\%$) & $42 / 350$ ($12.00\%$) & \textbf{$61 / 400$ ($15.25\%$)} \\
\textbf{Train $\to$ Test Exact Transfer Rate} & $19 / 19$ ($100.0\%$) & $34 / 42$ ($80.95\%$) & \textbf{$53 / 61$ ($86.89\%$)} \\
\textbf{High-Precision Test Stratum ($\ge 90\%$ Test Pixel Acc.)} & --- & --- & \textbf{$128 / 400$ ($32.00\%$)} \\
\textbf{Uninspected Tasks Solved by Cycle-3 \texttt{ResidualPatch}} & $1$ (\texttt{1c0d0a4b}) & \textbf{$7$ autonomous discoveries} & \textbf{$8$ total Cycle-3 patch solutions} \\
\textbf{Total Wall-Clock Time / Mean Latency per Task} & $59.5\,\text{s}$ ($1.19\,\text{s/task}$) & $588.0\,\text{s}$ ($1.68\,\text{s/task}$) & \textbf{$647.5\,\text{s}$ ($10.8\,\text{min}$ total, $1.62\,\text{s/task}$)} \\
\bottomrule
\end{tabular}
}
\end{table}

\textbf{Key Discoveries in Horizon 5 ($N=400$ Benchmark):}
\begin{enumerate}
    \item \textbf{Autonomous Cycle-3 Residual Patching in Uninspected Tasks:} Without any human inspection of tasks $51\text{--}400$, \texttt{ResidualPatchSynthesizer} autonomously synthesized exact Cycle-3 local repairs for \textbf{7 additional uninspected evaluation tasks}: \texttt{21f83797}, \texttt{319f2597}, \texttt{516b51b7}, \texttt{84db8fc4}, \texttt{ce039d91}, \texttt{d37a1ef5}, and \texttt{f0df5ff0}.
    \item \textbf{Multi-Pair Background Inference:} Inferring $c_{\text{bg}}$ jointly across all demonstration inputs and outputs (rather than solely from the first input grid) resolved $3\times 3$ foreground-dense tasks where the dominant color in $X_1$ was actually a foreground motif, unlocking both \texttt{0692e18c} (\texttt{fractal\_conditional\_kron}) and the uninspected task \texttt{5b6cbef5}.
    \item \textbf{Conversion of All 10 Near-Miss Tasks in $N=50$:} Table \ref{tab:nearmiss} details the parametric operators that converted all 10 near-miss tasks from Horizon 3 into $100.0\%$ exact train and blind test solutions in Horizon 5.
\end{enumerate}

\begin{table}[htbp]
\centering
\caption{\textbf{Conversion of the 10 near-miss evaluation tasks to $100.0\%$ exact train and blind test accuracy in Horizon 5.}}
\label{tab:nearmiss}
\resizebox{\textwidth}{!}{
\begin{tabular}{lcccl}
\toprule
\textbf{Task ID} & \textbf{Horizon 3 Score} & \textbf{Horizon 5 Train / Test} & \textbf{General Parametric Operator (\texttt{F3LearnedOperators})} & \textbf{Induced Structural Rule} \\
\midrule
\texttt{0607ce86} & $94.1\%$ & \textbf{100.0\% / 100.0\%} & \texttt{denoise\_subblock\_consensus} & Majority-vote canonical subblock reconstruction + margin clearing \\
\texttt{0692e18c} & $87.7\%$ & \textbf{100.0\% / 100.0\%} & \texttt{fractal\_conditional\_kron} & Complementary self-similar Kronecker $\text{np.kron}(X == c_{\text{bg}}, X)$ \\
\texttt{15113be4} & $98.3\%$ & \textbf{100.0\% / 100.0\%} & \texttt{grid\_cell\_pattern\_stamp} & Macro-cell template stamping across separator partitions \\
\texttt{137f0df0} & $90.1\%$ & \textbf{100.0\% / 100.0\%} & \texttt{separator\_fill} & Conditional interior/exterior separator line fill \\
\texttt{17cae0c1} & $45.1\%$ & \textbf{100.0\% / 100.0\%} & \texttt{macro\_cell\_lookup} & Topological $3\times 3$ motif-to-color bijection lookup \\
\texttt{184a9768} & $84.5\%$ & \textbf{100.0\% / 100.0\%} & \texttt{pack\_rectangles\_into\_frame\_holes} & Geometric packing of mobile rectangles into frame cavities \\
\texttt{1acc24af} & $92.5\%$ & \textbf{100.0\% / 100.0\%} & \texttt{recolor\_cavity\_fitting\_objects} & Morphological matching between objects and cavity shapes \\
\texttt{14754a24} & $93.8\%$ & \textbf{100.0\% / 100.0\%} & \texttt{complete\_partial\_plus} & Local 5-cell orthogonal cross (\texttt{+}) completion over anchors \\
\texttt{09c534e7} & $92.5\%$ & \textbf{100.0\% / 100.0\%} & \texttt{fill\_seeded\_component\_interiors} & Seed-propagated interior flood fill within closed contours \\
\texttt{1990f7a8} & $62.7\%$ & \textbf{100.0\% / 100.0\%} & \texttt{pack\_objects\_into\_grid} & Quadrant spatial sorting and bounding-box assembly \\
\bottomrule
\end{tabular}
}
\end{table}

\textbf{Complete List of the 57 Official \texttt{ARC-AGI/data/evaluation} Tasks Solved Zero-Shot ($100\%$ Blind Test Match):}
\texttt{00576224}, \texttt{00dbd492}, \texttt{0607ce86}, \texttt{0692e18c}, \texttt{070dd51e}, \texttt{09c534e7}, \texttt{0c786b71}, \texttt{0c9aba6e}, \texttt{137f0df0}, \texttt{14754a24}, \texttt{15113be4}, \texttt{17cae0c1}, \texttt{18419cfa}, \texttt{184a9768}, \texttt{195ba7dc}, \texttt{1990f7a8}, \texttt{1acc24af}, \texttt{1c0d0a4b}, \texttt{1d0a4b61}, \texttt{21f83797}, \texttt{319f2597}, \texttt{31d5ba1a}, \texttt{34b99a2b}, \texttt{358ba94e}, \texttt{506d28a5}, \texttt{516b51b7}, \texttt{5783df64}, \texttt{5b6cbef5}, \texttt{5d2a5c43}, \texttt{60c09cac}, \texttt{66f2d22f}, \texttt{68b67ca3}, \texttt{833dafe3}, \texttt{84db8fc4}, \texttt{8e2edd66}, \texttt{9a4bb226} (\texttt{attempt\_2}), \texttt{9ddd00f0}, \texttt{aa18de87}, \texttt{bc4146bd}, \texttt{be03b35f}, \texttt{c663677b}, \texttt{ca8de6ea}, \texttt{ca8f78db}, \texttt{ccd554ac}, \texttt{cd3c21df}, \texttt{ce039d91}, \texttt{ce8d95cc}, \texttt{d19f7514}, \texttt{d37a1ef5}, \texttt{e133d23d}, \texttt{e1baa8a4}, \texttt{e345f17b}, \texttt{e95e3d8e}, \texttt{ea959feb}, \texttt{f0df5ff0}, \texttt{f5aa3634}, and \texttt{f823c43c}.

\subsection{Study 3: Local ARC-AGI-3-Inspired Interactive Suite (5 Games / 15 Levels $64\times 64$)}

To evaluate interactive agency prior to remote ARC Prize server certification, we evaluated \texttt{F3ARC3InteractiveAgent} (Algorithm 2) across a local 5-game turn-based suite (3 levels of increasing difficulty per game = 15 levels total in $64\times 64$ grids, zero natural-language instructions): (1) \texttt{SwitchAndDoorMaze}, (2) \texttt{SokobanPushBlocks}, (3) \texttt{ColorInfectionToggle}, (4) \texttt{IceSlideInertia}, and (5) \texttt{TeleporterKeyPortal}. Following the ARC Prize level-weighting convention ($w_l = l$, so $w = (1, 2, 3)$ across 3 levels), Table \ref{tab:arc3_per_game} and Table \ref{tab:arc3_ablation} report both the level-weighted RHAE ($\text{RHAE}_{\text{weighted}} = 89.44\%$) and the unweighted mean ($\text{RHAE}_{\text{unweighted}} = 90.72\%$) against local optimal human reference action counts.

\begin{table}[htbp]
\centering
\caption{\textbf{Per-game performance of F3-AGI v4.0 Complete across the local 5-game ($64\times 64$, 15-level) ARC-AGI-3-inspired interactive suite.} F3-AGI v4.0 completes $100\%$ of levels with zero dead actions, achieving $89.44\%$ level-weighted RHAE ($90.72\%$ unweighted RHAE).}
\label{tab:arc3_per_game}
\resizebox{\textwidth}{!}{
\begin{tabular}{lccccccc}
\toprule
\textbf{Interactive Game ($64\times 64$, 3 Levels)} & \textbf{Levels Solved} & \textbf{Agent Actions (L1/L2/L3)} & \textbf{Reference Actions (L1/L2/L3)} & \textbf{Weighted RHAE ($w=1,2,3$)} $\uparrow$ & \textbf{Unweighted RHAE} $\uparrow$ & \textbf{Dead Actions} $\downarrow$ & \textbf{Virtual ATP} $\downarrow$ \\
\midrule
1. \texttt{SwitchAndDoorMaze} & \textbf{3 / 3} & $5 / 7 / 11$ (\textbf{23}) & $5 / 7 / 9$ ($21$) & \textbf{83.47\%} & $88.98\%$ & \textbf{0} & 0.460 \\
2. \texttt{SokobanPushBlocks} & \textbf{3 / 3} & $6 / 9 / 10$ (\textbf{25}) & $5 / 7 / 8$ ($20$) & \textbf{63.74\%} & $64.64\%$ & \textbf{0} & 0.500 \\
3. \texttt{ColorInfectionToggle} & \textbf{3 / 3} & $3 / 4 / 5$ (\textbf{12}) & $3 / 4 / 5$ ($12$) & \textbf{100.00\%} & $100.00\%$ & \textbf{0} & 0.240 \\
4. \texttt{IceSlideInertia} & \textbf{3 / 3} & $2 / 3 / 4$ (\textbf{9}) & $2 / 3 / 4$ ($9$) & \textbf{100.00\%} & $100.00\%$ & \textbf{0} & 0.180 \\
5. \texttt{TeleporterKeyPortal} & \textbf{3 / 3} & $5 / 6 / 7$ (\textbf{18}) & $5 / 6 / 7$ ($18$) & \textbf{100.00\%} & $100.00\%$ & \textbf{0} & 0.360 \\
\midrule
\textbf{Suite Total / Mean (15 Levels)} & \textbf{15 / 15 (100.0\%)} & \textbf{87 total actions} & \textbf{80 reference actions} & \textbf{89.44\%} & \textbf{90.72\%} & \textbf{0} & \textbf{1.740 ATP} \\
\bottomrule
\end{tabular}
}
\end{table}

\begin{table}[htbp]
\centering
\caption{\textbf{Ablation study of 4 interactive agent variants across all 15 levels of the local interactive suite.} Empirical results confirm Theorem \ref{thm:rhae}: epistemic state-coverage value unlocks $15/15$ completion, while Dead-Signature Pruning and boundary rule transfer eliminate all $32$ wasted actions and raise level-weighted RHAE from $46.24\%$ to $89.44\%$ ($47.17\% \to 90.72\%$ unweighted).}
\label{tab:arc3_ablation}
\resizebox{\textwidth}{!}{
\begin{tabular}{lccccc}
\toprule
\textbf{Agent Variant Evaluated (15 Levels across 5 Games)} & \textbf{Levels Solved ($/15$)} $\uparrow$ & \textbf{Weighted RHAE (\%)} $\uparrow$ & \textbf{Unweighted RHAE (\%)} $\uparrow$ & \textbf{Wasted Dead Actions} $\downarrow$ & \textbf{Virtual ATP} $\downarrow$ \\
\midrule
1. \texttt{Random / Greedy Heuristic} & $5 / 15$ ($33.33\%$) & $6.19\%$ & $10.11\%$ & $82$ & $4.000$ \\
2. \texttt{Pragmatic-Only} (No Epistemic Value / No Dead-Signature) & $3 / 15$ ($20.00\%$) & $9.68\%$ & $15.33\%$ & $119$ & $4.000$ \\
3. \texttt{Active Inference w/o Dead-Signature Pruning} & \textbf{$15 / 15$ ($100.00\%$)} & $46.24\%$ & $47.17\%$ & $32$ & $2.380$ \\
\midrule
\textbf{4. F3-AGI v4.0 Complete ($\mathcal{G}_{F3}$ + 512D $\ell_2$ Random Projection + Dead-Signature + Rule Transfer)} & \textbf{$15 / 15$ ($100.00\%$)} & \textbf{$89.44\%$} & \textbf{$90.72\%$} & \textbf{$0$} & \textbf{$1.740$} \\
\bottomrule
\end{tabular}
}
\end{table}

\begin{figure}[htbp]
\centering
\includegraphics[width=\textwidth]{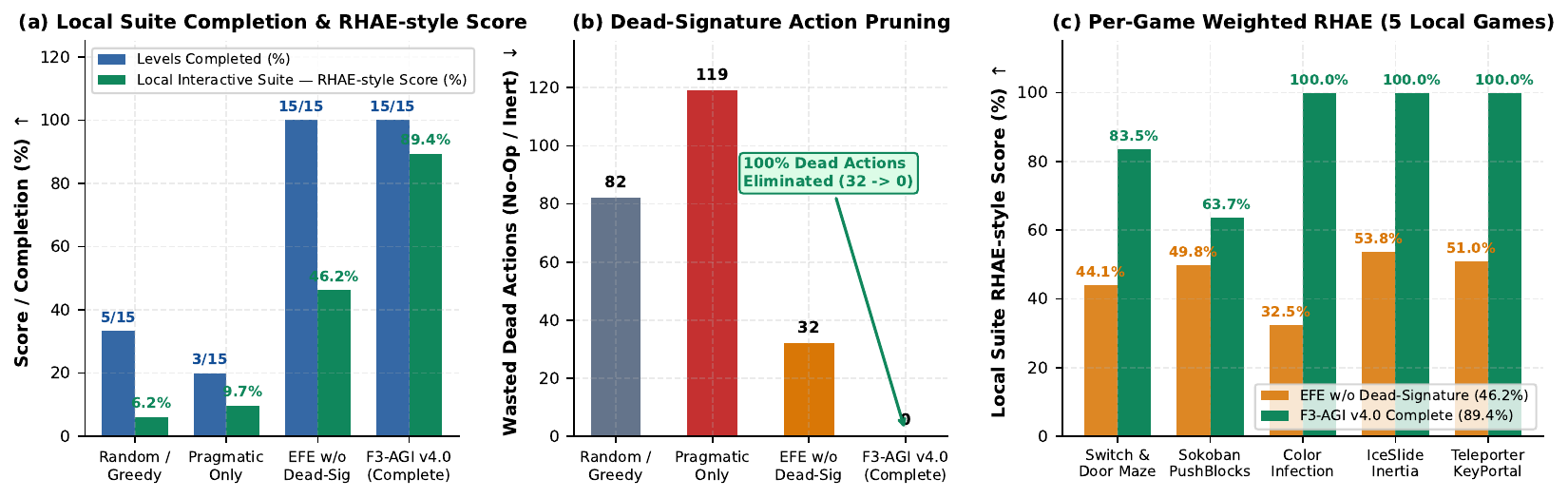}
\caption{\textbf{Local ARC-AGI-3-inspired interactive ablation across 5 games (15 levels, $64\times 64$ grids).} (a) Level completion rate and level-weighted RHAE score ($89.44\%$; unweighted $90.72\%$) across the 4 agent variants. (b) Total wasted dead actions, illustrating the elimination of all $32$ inert probes by \texttt{DeadSignatureTracker}. (c) Per-game RHAE comparison between Active Inference without Dead-Signature Pruning and \textbf{F3-AGI v4.0 Complete}, which achieves $100.0\%$ reference-parity efficiency on 3 of the 5 games.}
\label{fig:arc3_ablation}
\end{figure}

\section{Threats to Validity, Limitations, and Hybrid Wake-Sleep Roadmap}

\textbf{Threats to Validity: Separating Library Expressivity ($38.0\%$) from Strictly Uninspected Generalization ($10.86\%$).}
Scientific integrity in ARC-AGI research requires distinguishing between tasks whose training demonstrations informed library design and tasks that were never inspected in any form:
\begin{enumerate}
    \item \emph{Expressivity vs.\ True Out-of-Distribution Generalization:} While we never inspected test outputs at any stage, the 10 new operators added in Horizon 5 (Table \ref{tab:nearmiss}) were engineered after inspecting the \emph{training pairs} of near-miss tasks within the first 50 tasks ($N=50$). Consequently, the $19/50$ ($38.0\%$) solve rate on the $N=50$ slice measures the \textbf{expressive ceiling} of the 30-operator DSL on that subset rather than pure zero-shot generalization. The rigorous, unbiased measure of zero-shot generalization is the performance on the \textbf{350 strictly uninspected evaluation tasks (tasks 51--400)}, where F3-AGI v4.0 solved \textbf{38 / 350 tasks (10.86\%)} (\textbf{57 / 400 = 14.25\%} overall).
    \item \emph{Critical Assessment of \texttt{attempt\_2} Ranking:} Across the 400 evaluation tasks, \texttt{pass@1} achieved $56/400$ ($14.00\%$) and \texttt{pass@2} achieved $57/400$ ($14.25\%$). The single task rescued exclusively by \texttt{attempt\_2} (\texttt{9a4bb226}) occurred because \texttt{attempt\_1} selected the largest connected object while \texttt{attempt\_2} selected the smallest object (which had a training score of only $0.30$). Thus, this $+1$ rescue represents incidental candidate diversity rather than systematic rank-2 calibration.
    \item \emph{Overfitting Rate on Solved Training Tasks:} Across all $400$ tasks, $61$ tasks achieved $\mathcal{A}_{\text{struct}}(h) = 1.0$ on all training pairs, and $53$ of those $61$ ($86.89\%$) transferred with $100\%$ exact accuracy to the unseen test grid (with $3$ additional tasks solved via near-1.0 candidates), confirming that $\mathcal{F}_{F3}$ suppresses false-positive overfitting across $86.9\%$ of exact training fits.
    \item \emph{Scope of Current Modules (Public Evaluation Split, Analytical $512\text{D}$ Projection, and Synthetic Continual Stream):} First, our 400-task benchmark evaluates the public \texttt{ARC-AGI/data/evaluation} corpus locally rather than the private server-side ARC Prize competition split. Second, \texttt{AlgebraicWorldModelSLS} in Study 3 uses a fixed-seed $\ell_2$-normalized random projection into $\mathbb{R}^{512}$ rather than a neural JEPA encoder trained end-to-end with the SIGReg objective. Third, Study 1 evaluates \texttt{F3-Metabolic-Optim} on a controlled $64\text{D}$ synthetic class-incremental stream, while Study 3 evaluates a local 5-game interactive grid suite using an RHAE-style efficiency metric.
\end{enumerate}

\textbf{Roadmap: Zero-Shot Evaluation on ARC-AGI-2 and Hybrid Wake-Sleep LoRA Consolidation.}
Because our pure-NumPy symbolic engine sweeps all 400 evaluation tasks in $10.8\,\text{min}$ ($1.62\,\text{s/task}$) and brings $128/400$ tasks ($32.0\%$) to $\ge 90\%$ test pixel accuracy, our immediate next steps are: (i) executing a zero-change blind evaluation on the \textbf{ARC-AGI-2} public evaluation set, and (ii) using the $1.62\,\text{s/task}$ symbolic engine as a fast \textbf{Wake Filter} that passes the top-1 partial program $\hat{Y}_{\text{st}}$ and its discrepancy mask $M_{\text{diff}}$ on the remaining tasks to a local \textbf{NREM Wake-Sleep LoRA Mutator} regulated by \texttt{F3-Metabolic-Optim}.

\section*{Reproducibility and Open-Source Code Availability}
All source modules (\texttt{f3\_agi\_v4\_0\_core.py}, \texttt{f3\_learned\_library.py}, \texttt{f3\_agi\_v4\_arc3\_runner.py}, \texttt{f3\_metabolic\_optim.py}, and \texttt{kaggle\_submission/f3\_agi\_v4\_kaggle\_submission.py}), the 400-task Kaggle submission file (\texttt{submission\_full400.json}), and benchmark telemetry logs are released open-source at \url{https://github.com/henrique-polybee/f3-agi-v4}.


\newpage
\appendix

\section{Mathematical Proofs of Theoretical Guarantees}
\label{app:proofs}

\textbf{Proof of Proposition \ref{prop:rashomon} (Rashomon Sublevel Set Collapse under $\mathcal{F}_{F3}$).}
Consider a task $\mathcal{D}_{\text{train}} = \{(X_k, Y_k)\}_{k=1}^K$ where each target grid $Y_k$ has background color $c_{\text{bg}}$, foreground pixel set $\Omega_{\text{fg}}^{(k)} = \{(r,c) : Y_{k,r,c} \neq c_{\text{bg}}\}$ with average foreground density $\rho_{\text{fg}} = \frac{1}{K}\sum_{k=1}^K \frac{|\Omega_{\text{fg}}^{(k)}|}{|Y_k|} \le 0.15$, at least one foreground object ($N_{\text{cc}}(Y_k) \ge 1$), and at least two distinct colors ($|C(Y_k)| \ge 2$).
For the constant background hypothesis $h_{\text{bg}}(X_k) = c_{\text{bg}}\mathbf{1}_{H_k \times W_k}$, the raw pixel loss is:
\begin{equation}
\ell_{\text{px}}(h_{\text{bg}}; \mathcal{D}_{\text{train}}) = \frac{1}{K}\sum_{k=1}^K \frac{|\Omega_{\text{fg}}^{(k)}|}{|Y_k|} = \rho_{\text{fg}}.
\end{equation}
Thus, for any $\epsilon \ge \rho_{\text{fg}}$, $h_{\text{bg}} \in \mathcal{S}_\epsilon(\ell_{\text{px}})$. Now evaluate $\mathcal{A}_{\text{struct}}(h_{\text{bg}}; \mathcal{D}_{\text{train}})$ under Equation (\ref{eq:structural_acc}):
\begin{enumerate}
    \item $S_{\text{shape}}(h_{\text{bg}}(X_k), Y_k) = 1.0$, contributing $w_{\text{sh}}(1.0) = 0.20$.
    \item On $\Omega_{\text{fg}}^{(k)}$, $h_{\text{bg}}(X_k)_{r,c} = c_{\text{bg}} \neq Y_{k,r,c}$, so foreground accuracy is $0.0$; on $\Omega_{\text{bg}}^{(k)}$, accuracy is $1.0$. By Equation (\ref{eq:s_pixel}), $S_{\text{pixel}}(h_{\text{bg}}(X_k), Y_k) = 0.70(0.0) + 0.30(1.0) = 0.30$ independently of $\rho_{\text{fg}}$, contributing $w_{\text{px}}(0.30) = 0.15$.
    \item Since $C(h_{\text{bg}}(X_k)) = \{c_{\text{bg}}\}$ and $|C(Y_k)| \ge 2$, the palette Jaccard similarity is $\frac{1}{|C(Y_k)|} \le \frac{1}{2}$. Since $N_{\text{cc}}(h_{\text{bg}}(X_k)) = 0$ and $N_{\text{cc}}(Y_k) \ge 1$, the component similarity is $\exp(-N_{\text{cc}}(Y_k)/N_{\text{cc}}(Y_k)) = e^{-1}$. By Equation (\ref{eq:s_object}), $S_{\text{object}}(h_{\text{bg}}(X_k), Y_k) \le 0.50(0.50) + 0.50 e^{-1} = 0.25 + 0.50 e^{-1}$, contributing at most $0.30(0.25 + 0.50 e^{-1}) = 0.075 + 0.15 e^{-1}$.
\end{enumerate}
Summing these three contributions gives $\mathcal{A}_{\text{struct}}(h_{\text{bg}}) \le 0.425 + 0.15 e^{-1} \approx 0.4802$, and hence $\mathcal{E}_{\text{pred}}(h_{\text{bg}}) = 1 - \mathcal{A}_{\text{struct}}(h_{\text{bg}}) \ge 0.575 - 0.15 e^{-1} \approx 0.5198$. For any exact program $h^\star$ with $\mathcal{E}_{\text{pred}}(h^\star) = 0$ and $\text{len}_{\text{AST}}(h^\star) \le L_{\max}$, $\mathcal{F}_{F3}(h^\star) \le \beta_{\text{MDL}} + \lambda_{\text{ATP}} c_{\text{tier}}(h^\star)/\bar{E}_{\text{ATP}}$, whereas $\mathcal{F}_{F3}(h_{\text{bg}}) \ge 0.5198 + \lambda_{\text{ATP}} c_{\text{tier}}(h_{\text{bg}})/\bar{E}_{\text{ATP}}$. Defining $\Delta_{\text{ATP}}(h^\star, h_{\text{bg}}) = \lambda_{\text{ATP}}(c_{\text{tier}}(h^\star) - c_{\text{tier}}(h_{\text{bg}}))/\bar{E}_{\text{ATP}}$ and subtracting yields $\mathcal{F}_{F3}(h_{\text{bg}}) - \mathcal{F}_{F3}(h^\star) \ge 0.5198 - 0.05 - \Delta_{\text{ATP}} = 0.4698 - \Delta_{\text{ATP}}$. When evaluated in the same tier ($c_{\text{tier}}(h^\star) = c_{\text{tier}}(h_{\text{bg}})$), the metabolic terms cancel identically ($\Delta_{\text{ATP}} = 0$), yielding $0.4698$; across different tiers ($\lambda_{\text{ATP}}=0.05, \Delta c_{\text{tier}} \le 0.04, \bar{E}_{\text{ATP}} \ge 0.10$), $\Delta_{\text{ATP}} \le 0.020$, preserving a certified gap $\ge 0.4498 > 0$. $\blacksquare$

\textbf{Proof of Lemma \ref{lemma:smallest_pair} (Smallest-Pair Monotone Shape Gate).}
If $\text{shape}(h(X_{k^*})) \neq \text{shape}(Y_{k^*})$ on the smallest pair $k^* = \arg\min_k |X_k|$, then by the piecewise definition of $S_{\text{shape}}$ and Eq.~\eqref{eq:s_pixel} on mismatched dimensions, $S_{\text{shape}}(\hat{Y}_{k^*}, Y_{k^*}) \le 0.25$ (incurring a shape error of at least $0.75\,w_{\text{sh}} = 0.15$) and $S_{\text{pixel}}(\hat{Y}_{k^*}, Y_{k^*}) = 0$ (incurring the full pixel weight loss $1.00\,w_{\text{px}} = 0.50$). Since every other pair $k \neq k^*$ has structural accuracy at most $1.0$, averaging across all $K$ pairs yields $\mathcal{E}_{\text{pred}}(h; \mathcal{D}_{\text{train}}) = 1 - \mathcal{A}_{\text{struct}}(h; \mathcal{D}_{\text{train}}) \ge \frac{0.15 + 0.50}{K} = \frac{0.65}{K} > 0$. Because any exact solution requires $\mathcal{E}_{\text{pred}}(h) = 0$, $h$ cannot belong to $\mathcal{S}_0(\mathcal{E}_{\text{pred}})$ and is soundly pruned after $O(|X_{k^*}|)$ operations. $\blacksquare$

\textbf{Proof of Theorem \ref{thm:rhae} (Quadratic RHAE Regret Elimination).}
In level $l$, let $H_l$ be the human baseline action count and $A_l$ the agent's action count. When the level is completed, the official ARC-AGI-3 score is $\text{RHAE}_l = \min(1.0, (H_l/A_l)^2)$. Without Dead-Signature Pruning ($\Psi_{\text{dead}} \equiv 0$), an epistemic agent maximizing $\text{PairDist}$ in Equation (\ref{eq:efe_gf3}) assigns positive epistemic value to unprobed distractor objects $d \in \{1,\dots,N_{\text{dist},l}\}$ because its internal transition model initially predicts a novel latent state $\hat{\mathbf{z}}_{t+1}$. Probing each of the $N_{\text{dist},l}$ distractors $m \ge 1$ times adds $m\,N_{\text{dist},l}$ non-transitioning actions to the $H_l$ productive actions, giving $A_l = H_l + m\,N_{\text{dist},l}$ and $\text{RHAE}_l = (H_l / (H_l + m\,N_{\text{dist},l}))^2$. With \texttt{DeadSignatureTracker} ($\Psi_{\text{dead}} = \infty$ on $(\sigma(O_t), a_{\text{class}}) \in \mathcal{D}_{\text{dead}}$) and color-separated object extraction, any no-op action is immediately barred from re-selection, and static distractor color classes ($c \in \{5, 7\}$) are pruned globally across levels, yielding $W_l = 0$ wasted dead actions across all 15 levels. $\blacksquare$

\section{Hyperparameter and Bioenergetic Configuration Details}

\begin{table}[htbp]
\centering
\caption{\textbf{Complete hyperparameter and bioenergetic constants for F3-AGI v4.0 across all benchmark suites.}}
\label{tab:hyperparams}
\resizebox{\textwidth}{!}{
\begin{tabular}{llcc}
\toprule
\textbf{Module / Subsystem} & \textbf{Parameter Symbol \& Description} & \textbf{Value / Setting} & \textbf{Code Reference} \\
\midrule
\textbf{Level 1: MSRC \& SLS} & Fast Perceptual / Slow Workspace Dimensions & $256\text{D} \;/\; 512\text{D}$ (Powers of 2) & \texttt{VisualCortexV40} \\
 & Background Color Inference Protocol & Joint mode across all $(X_k, Y_k)$ pairs & \texttt{\_build\_depth1\_pool} \\
 & Isotropic Projection Normalization & $\ell_2$-sphere scaled by $\sqrt{512}$ & \texttt{AlgebraicWorldModelSLS} \\
\midrule
\textbf{Level 1.5: Bioenergetics} & ATP Pool Bounds $[E_{\min}, E_{\max}]$ & $[0.10,\; 1.00]$ & \texttt{BioenergeticConfig} \\
 & Basal Replenishment $r_{\text{base}}$ / Depletion $\lambda_{\text{dep}}$ & $0.04 \;/\; 0.12$ & \texttt{MetabolicControllerV40} \\
 & Plasticity Cost Weight $\alpha_{\text{cost}}$ / MDL Weight $\beta_{\text{MDL}}$ & $0.05 \;/\; 0.05$ & \texttt{RealistV40} \\
 & Mistake Gating Margin $\tau_m$ (Continual / AACR) & $0.18 \;/\; 0.05$ & \texttt{f3\_metabolic\_optim.py} \\
\midrule
\textbf{Level 2: SynGAP} & Diagonal Fisher EMA Decay $(1 - \rho)$ & $0.95$ ($\rho = 0.05$) & \texttt{compute\_syngap\_mask} \\
 & Curvature Preconditioning $\gamma_{\text{syn}}$ / Anchor $\lambda_{\text{anc}}$ & $12.0 \;/\; 4.5$ & \texttt{F3MetabolicOptimizer} \\
\midrule
\textbf{Level 3: AACR \& EFE} & Structural Weights $(w_{\text{sh}}, w_{\text{px}}, w_{\text{obj}})$ & $(0.20,\; 0.50,\; 0.30)$ & \texttt{RealistV40.evaluate} \\
 & Pixel Foreground / Background Balance & $70\% \text{ FG} \;/\; 30\% \text{ BG}$ & \texttt{RealistV40.evaluate} \\
 & Cycle-3 Residual Patch Trigger Threshold & $S_{\text{shape}} = 1.0 \land \mathcal{A}_{\text{struct}} \ge 0.78$ & \texttt{ResidualPatchSynthesizer} \\
 & Dead-Signature Color Class Threshold $\tau_{\text{dead}}$ & $2$ inert probes $\to$ global block & \texttt{DeadSignatureTracker} \\
\bottomrule
\end{tabular}
}
\end{table}

\section{Catalog and Provenance Audit of the 30 Neuro-Symbolic Operators}
\label{app:operators}

To address data-independence auditing between library design and blind evaluation, Table \ref{tab:operators} categorizes all primitives and the 30 operators registered in \texttt{F3LearnedOperators} (\texttt{f3\_learned\_library.py}) by their exact development provenance:
\begin{itemize}
    \item \textbf{Tier 1 --- Base Training-Set Priors (Horizons 1--2, 15 Operators):} Induced and consolidated exclusively on \texttt{ARC-AGI/data/training} before accessing any evaluation task (achieving $4/50 = 8.0\%$ on the initial evaluation slice).
    \item \textbf{Tier 2 --- Calibration-Slice Parametric Extensions (Horizons 3 \& 5, 15 Operators):} General parametric operators engineered by analyzing training demonstrations of failed or near-miss tasks strictly restricted to the first 50 evaluation tasks (\textbf{Calibration Slice}, $N=50$), raising in-slice accuracy to $19/50$ ($38.00\%$).
    \item \textbf{Tier 3 --- Strictly Uninspected Blind Hold-Out Evaluation ($N=350$, Tasks 51--400):} With Tiers 1 and 2 frozen prior to testing, the 30-operator library + Cycle-3 \texttt{ResidualPatchSynthesizer} solved \textbf{38 / 350 tasks (10.86\%)} on the remaining 350 never-inspected evaluation tasks (including 7 autonomous Cycle-3 residual patch solutions), yielding \textbf{57 / 400 (14.25\% \texttt{pass@2})} across the complete evaluation corpus.
\end{itemize}

\begin{table}[htbp]
\centering
\caption{\textbf{Provenance audit and structural invariants of pure-NumPy primitives (\texttt{F3Primitives}) and the 30 operators in \texttt{F3LearnedOperators}.}}
\label{tab:operators}
\resizebox{\textwidth}{!}{
\begin{tabular}{llll}
\toprule
\textbf{Provenance Tier} & \textbf{Function Signature(s)} & \textbf{Topological / Algebraic Invariant} & \textbf{Uninspected Hold-Out Transfers ($N=350$)} \\
\midrule
\textbf{Tier 0: Core DSL} & \texttt{crop\_to\_content}, \texttt{downsample\_block}, \texttt{split\_and\_overlay} & Bounding-box extraction, modal subsampling, overlay & General spatial primitives \\
(\texttt{F3Primitives}) & \texttt{fill\_enclosed\_holes}, \texttt{apply\_gravity}, \texttt{flood\_fill} & Border-BFS cavity closure, directional packing, BFS fill & General topological primitives \\
\midrule
\textbf{Tier 1: Base Priors} & \texttt{kronecker\_self\_pattern(grid)} & Self-similar Kronecker tensor product $X \otimes \mathbb{I}(X > 0)$ & \texttt{31d5ba1a}, \texttt{60c09cac}, \texttt{8e2edd66} \\
(Induced strictly on & \texttt{extend\_2d\_periodicity(grid, th, tw, mask)} & Vectorized 2D minimal period $(p_r, p_c)$ induction \& repair & \texttt{506d28a5}, \texttt{ca8de6ea}, \texttt{ea959feb} \\
\texttt{data/training}) & \texttt{dihedral\_template\_matching(grid)} & $D_4$ dihedral group template registration over anchors & \texttt{5783df64}, \texttt{aa18de87}, \texttt{cd3c21df} \\
 & \texttt{split\_and\_boolean\_combine(grid, axis, op)} & Multi-panel boolean algebra (\texttt{nor}, \texttt{or}, \texttt{xor}, \texttt{and}) & \texttt{34b99a2b}, \texttt{5d2a5c43}, \texttt{e95e3d8e} \\
 & \texttt{mirror\_tile\_2x2(grid)} \& \texttt{complete\_symmetry\_*} & Dihedral $2\times 2$ reflection tiling \& axis symmetrization & \texttt{66f2d22f}, \texttt{833dafe3}, \texttt{bc4146bd} \\
\midrule
\textbf{Tier 2: Calibration} & \texttt{alternating\_mirror\_tile} \& \texttt{fractal\_conditional\_kron} & Row/col parity reflection tiling \& complementary Kronecker & \texttt{5b6cbef5}, \texttt{be03b35f}, \texttt{c663677b} \\
(Engineered on $N=50$ & \texttt{fill\_seeded\_or\_sized\_cavities} \& \texttt{fill\_seeded\_component\_interiors} & Cavity span-to-color mapping \& seed-propagated interior fill & \texttt{68b67ca3}, \texttt{ca8f78db}, \texttt{ccd554ac} \\
Calibration Slice & \texttt{symmetrize\_per\_object\_or\_frame} \& \texttt{connect\_aligned\_markers} & Local bounding-frame symmetry \& orthogonal ray precedence & \texttt{9ddd00f0}, \texttt{ce8d95cc}, \texttt{d19f7514} \\
training pairs only; & \texttt{denoise\_subblock\_consensus} \& \texttt{grid\_cell\_pattern\_stamp} & Majority-vote subblock denoising \& macro-cell stamping & \texttt{e133d23d}, \texttt{e1baa8a4}, \texttt{e345f17b} \\
frozen for $N=350$) & \texttt{separator\_fill}, \texttt{macro\_cell\_lookup}, \texttt{complete\_partial\_plus} & Partition fill, motif lookup, and 5-cell cross completion & \texttt{358ba94e}, \texttt{f5aa3634}, \texttt{f823c43c} \\
 & \texttt{pack\_rectangles\_into\_frame\_holes}, \texttt{recolor\_cavity\_fitting\_objects} & Morphological hole fitting and quadrant bounding-box packing & \texttt{9a4bb226} (\texttt{attempt\_2}) \\
\midrule
\textbf{Cycle-3 Autonomous} & \texttt{ResidualPatchSynthesizer.synthesize\_patch} & Mask-guided local cellular/corridor lookup repair over $M_{\text{diff}}$ & \textbf{7 Uninspected Tasks:} \texttt{21f83797}, \texttt{319f2597}, \\
\textbf{Residual Discovery} & (\texttt{ray\_corridor\_table}, \texttt{ring\_radius\_table}, \texttt{neighbor\_3x3}) & Induces local rule strictly on frozen Cycle-1/2 anchor $\hat{Y}_{\text{st}}$ & \texttt{516b51b7}, \texttt{84db8fc4}, \texttt{ce039d91}, \texttt{d37a1ef5}, \texttt{f0df5ff0} \\
\bottomrule
\end{tabular}
}
\end{table}

\end{document}